\documentclass[10pt,twocolumn,letterpaper]{article}

\usepackage{wacv}              

\definecolor{wacvblue}{rgb}{0.21,0.49,0.74}
\usepackage[pagebackref,breaklinks,colorlinks,allcolors=wacvblue]{hyperref}
\usepackage{pifont}
\usepackage{xcolor}
\usepackage{bm}
\usepackage[accsupp]{axessibility}  

\def\wacvPaperID{121} 
\def\confName{WACV}
\def\confYear{2027}

\title{GUSH3R: Everyone Everywhere All at Once as Gaussians}

\author{Keito Abe \hspace{2em} Kaede Shiohara\thanks{Project lead.} \hspace{2em} Takashi Otonari \hspace{2em} Toshihiko Yamasaki\\
The University of Tokyo\\
{\tt\small \{abe, shiohara, otonari, yamasaki\}@cvm.t.u-tokyo.ac.jp}
}

\begin{document}
\maketitle

\begin{abstract}
Reconstructing dynamic human-scene environments from monocular videos is a challenging problem that requires jointly modeling scene geometry, camera motion, and non-rigid human dynamics while enabling photorealistic rendering. 
Recent feed-forward methods can efficiently predict geometry, but they are often limited to non-photorealistic representations such as point clouds and meshes, or they fail to handle non-rigid objects, particularly dynamic humans.
To fill this gap, we present \textbf{GUSH3R} (\textbf{G}aussian-\textbf{U}nified \textbf{S}cene \textbf{H}uman \textbf{3}D \textbf{R}econstruction), a feed-forward framework for online dynamic human-scene reconstruction.
From a monocular human-scene video, our method jointly reconstructs dynamic humans (\textit{everyone}) and static scenes (\textit{everywhere}) in a single forward pass per frame (\textit{all at once}) as 3D Gaussian Splatting primitives (\textit{as Gaussians}), which are geometrically consistent and capable of photorealistic novel view synthesis.
Experiments on monocular human-scene datasets demonstrate that our approach significantly improves inference efficiency over optimization-based methods while achieving competitive novel view synthesis quality against human-scene baselines. Moreover, GUSH3R outperforms general 4D reconstruction methods in novel view video synthesis, while operating in a feed-forward, pose-free, and streaming manner.
For implementation details, please refer to our source code at \url{https://github.com/abkeito/GUSH3R}.
\end{abstract}
  
\section{Introduction}
\label{sec: intro}
Reconstruction and novel view synthesis of dynamic human-scene environments from monocular videos are important for virtual and augmented reality, and digital human modeling~\cite{bullettime,AR-splatloc,guo2023vid2avatar}. 
Practical applications require both fast reconstruction and photorealistic rendering, which remain difficult to achieve jointly for monocular videos with dynamic humans.
\begin{figure}[t]
  \centering
  \includegraphics[width=\linewidth]{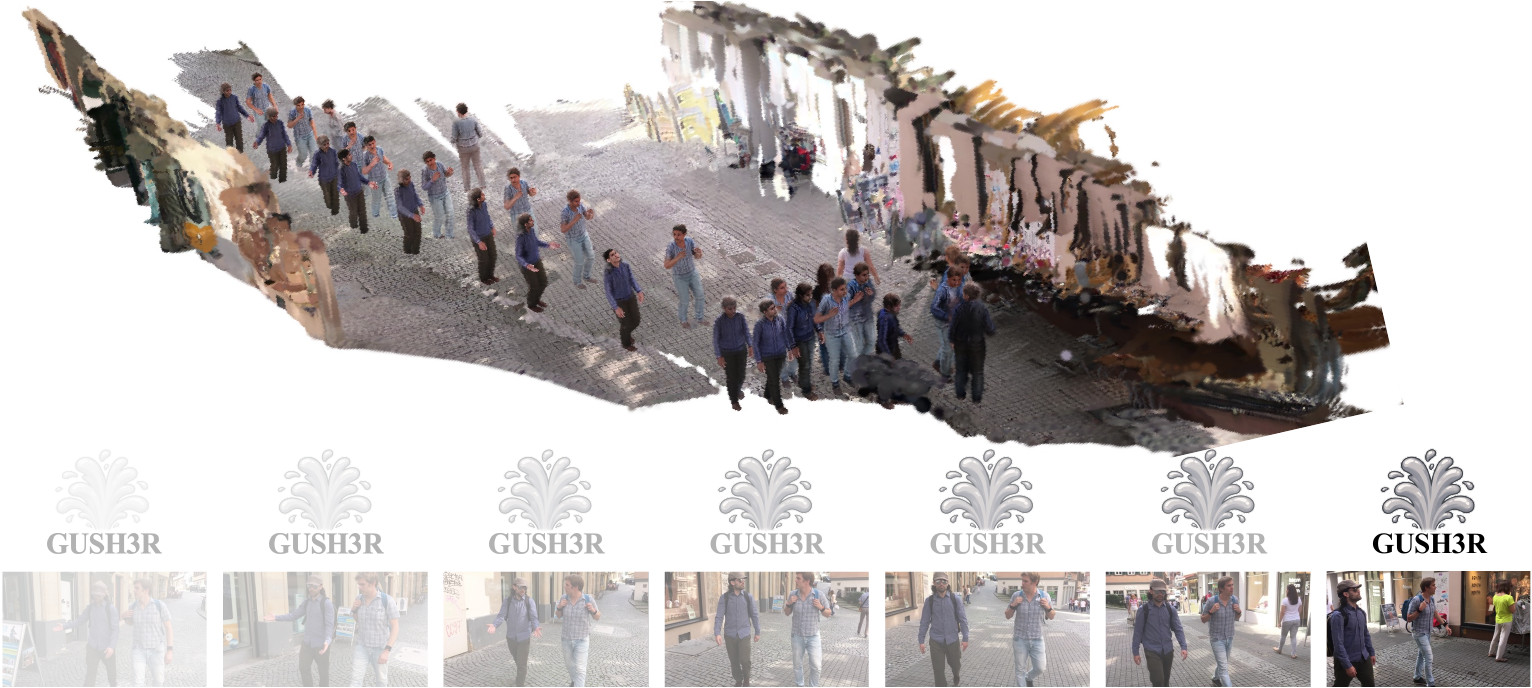}
  \caption{
  \textbf{GUSH3R} (\textbf{G}aussian-\textbf{U}nified \textbf{S}cene \textbf{H}uman \textbf{3}D \textbf{R}econstruction) takes a monocular video as input and produces dynamic human-scene representations using 3D Gaussians.
  }
  \label{fig:teaser}
\end{figure}

Optimization-based methods using Neural Radiance Fields (NeRF) and 3D Gaussian Splatting (3DGS) achieve high reconstruction quality but require costly per-scene optimization~\cite{mildenhall2020nerf,kerbl3Dgaussians,Wu_2024_CVPR-4dgs,dynamic-3dgaussians}, limiting rapid or interactive reconstruction~\cite{chen2024far-nerfmemory,xue2024hsr,zhang2025odhsr}. 
Moreover, several dynamic reconstruction methods rely on multi-view videos~\cite{Wu_2024_CVPR-4dgs,dynamic-3dgaussians,kplane}, limiting their applicability to scenarios where only a single video stream is available.

In contrast, feed-forward methods predict 3D geometry directly from images in a single forward pass, enabling fast inference~\cite{dust3r,mast3r,wang2025vggt,AnySplat,depthanything3,cut3r,zhangmonst3r}. 
These approaches generalize well to unseen scenes by leveraging strong geometric priors learned from large-scale data~\cite{co3dv2,blendedmvs,dl3dv,megadepth,kubric,wildrgb,scannet,hypersim,mapillary,habitat,replica,mvssynth,pointodyssey,virtualkitti,aria,objaverse}.
Yet, existing feed-forward methods still struggle to jointly handle dynamic humans and provide representations that support photorealistic rendering, leaving a key gap toward practical human-scene reconstruction applications~\cite{AnySplat,chen2025human3r}.

As summarized in Table~\ref{tab:concept}, existing feed-forward methods either do not explicitly model dynamic humans or do not provide representations suitable for photorealistic rendering.
In this work, we take a practical first step toward feed-forward dynamic human-scene reconstruction with photorealistic rendering from monocular videos and propose GUSH3R (Gaussian-Unified Scene Human 3D Reconstruction) as shown in Fig.~\ref{fig:teaser}.
To this end, we leverage geometric and human priors~\cite{chen2025human3r} and lift them into a unified 3DGS representation.
Our representation consists of dynamic human Gaussians and static scene Gaussians, whose appearance is predicted by respective decoders for humans and scenes.
The human decoder combines scene-aware visual features with articulated human priors to predict Gaussians that follow human motion, while the scene decoder predicts static Gaussians from geometric scene features.
Our method enables feed-forward reconstruction of dynamic human-scene environments while preserving the photorealistic rendering quality of a 3DGS representation.
Our contributions are as follows:
\begin{itemize}
    \item  To the best of our knowledge, we introduce the first method for \textit{feed-forward photorealistic, renderable dynamic human-scene reconstruction from monocular videos} that jointly models articulated humans and a static scene as 3D Gaussians through causal frame-by-frame processing, enabling rapid reconstruction without per-scene optimization.
    \item We design an architecture that bridges human-scene foundation models and photorealistic rendering by leveraging geometric priors and SMPL-X~\cite{SMPL-X} representations.
    \item We demonstrate that our method achieves competitive human-scene reconstruction quality with substantially improved efficiency, while also outperforming general 4D reconstruction baselines in novel view video synthesis under a feed-forward, pose-free, and streaming setting.
\end{itemize}

\begin{table}
\centering
\resizebox{\linewidth}{!}{
\begin{tabular}{lccc}
\toprule
Method & Streaming & Dynamic humans & Photo-reality \\
\midrule
VGGT~\cite{wang2025vggt} & {\color{red}\ding{55}}& {\color{red}\ding{55}}& {\color{red}\ding{55}} \\
AnySplat~\cite{AnySplat} & {\color{red}\ding{55}}& {\color{red}\ding{55}}& {\color{Green}\ding{51}}\\
CUT3R~\cite{cut3r} & {\color{Green}\ding{51}}& {\color{red}\ding{55}}& {\color{red}\ding{55}} \\
StreamSplat~\cite{wu2025streamsplat} & {\color{Green}\ding{51}}& {\color{red}\ding{55}}& {\color{Green}\ding{51}}\\
Human3R~\cite{chen2025human3r}  & {\color{Green}\ding{51}} & {\color{Green}\ding{51}}& {\color{red}\ding{55}}\\
Ours  & {\color{Green}\ding{51}} & {\color{Green}\ding{51}} & {\color{Green}\ding{51}} \\
\bottomrule
\end{tabular}
}

\caption{\textbf{Concept-level comparison of feed-forward models.} Here, ``Streaming'' denotes causal frame-by-frame processing without access to future frames, ``Dynamic humans'' indicates explicit modeling of non-rigid human motion, and ``Photo-reality'' refers to a renderable representation suitable for novel view synthesis.}
\label{tab:concept}
\end{table}
\section{Related Work}

\subsection{3D Reconstruction}
Early approaches to 3D reconstruction typically rely on multi-view geometry pipelines such as structure-from-motion (SfM)~\cite{rome,colmap,romecloudlessday,robustsfm} and multi-view stereo (MVS)~\cite{galliani2015massive-mvs,pixelwise}, which recover camera poses and explicit 3D structure from image correspondences. 
Neural rendering methods such as NeRF~\cite{mildenhall2020nerf} and Gaussian-based methods such as 3DGS~\cite{kerbl3Dgaussians} have also been introduced, which represent scenes using continuous or point-based representations optimized through differentiable rendering, enabling photorealistic novel view synthesis.
While these approaches produce high-quality reconstructions, they rely on iterative optimization over camera parameters and scene representations, making inference computationally expensive~\cite{li2023steernerf-quicknerf,chen2024far-nerfmemory}.

Recent reconstruction methods~\cite{dust3r, mast3r, mast3rsfm, wang2025vggt} incorporate feed-forward networks to predict scene geometry, including point maps and camera parameters, directly from input images, enabling fast inference and generalization to unseen scenes.
They achieve this by learning strong geometric priors from large-scale data~\cite{co3dv2,blendedmvs,dl3dv,megadepth,kubric,wildrgb,scannet,hypersim,mapillary,habitat,replica,mvssynth,pointodyssey,virtualkitti,aria,objaverse} and leveraging transformer-based architectures with the help of strong backbone features~\cite{vit,oquab2024dinov2}.
Feed-forward approaches such as AnySplat~\cite{AnySplat} and Depth Anything 3~\cite{depthanything3} further predict renderable 3DGS representations directly from images.

However, these feed-forward approaches primarily focus on static scenes and often struggle with dynamic objects, leading to inconsistent geometry and degraded reconstruction quality~\cite{AnySplat,depthanything3}. 
Our work addresses this limitation by explicitly disentangling dynamic humans from static scenes and modeling them separately within a unified 3DGS framework.

\subsection{4D Reconstruction}
Reconstructing dynamic scenes from image sequences has also been widely studied. 
Many existing approaches \cite{Wu_2024_CVPR-4dgs, pumarola2020d-dnerf, som2024,4dvisofdynamicevents} rely on optimization-based pipelines that model scene dynamics using deformation fields or canonical representations. 
They typically require multi-view synchronized video inputs~\cite{4dvisofdynamicevents,Wu_2024_CVPR-4dgs,cao2023hexplane,neural3dvidefrommultiview} and costly optimization, limiting their scalability and applicability in real-world scenarios.
More recently, feed-forward methods reconstruct dynamic scenes directly from monocular videos~\cite{cut3r,zhangmonst3r,chen2025ttt3r,streamVGGT,d4rt}.
Recent approaches further extend this paradigm to jointly estimate dynamic geometry and cameras~\cite{d4rt} or produce renderable representations~\cite{movies}.
Optimization-based methods~\cite{Wu_2024_CVPR-4dgs,som2024} require costly per-scene optimization, while feed-forward geometry reconstruction methods such as MonST3R and CUT3R~\cite{zhangmonst3r,cut3r} do not directly predict photorealistically renderable representations.
Our method jointly reconstructs articulated humans and a static background as 3D Gaussians in a shared coordinate system, processing monocular video frames sequentially without access to future frames.
\begin{figure*}[t]
  \centering
  \includegraphics[width=\linewidth]{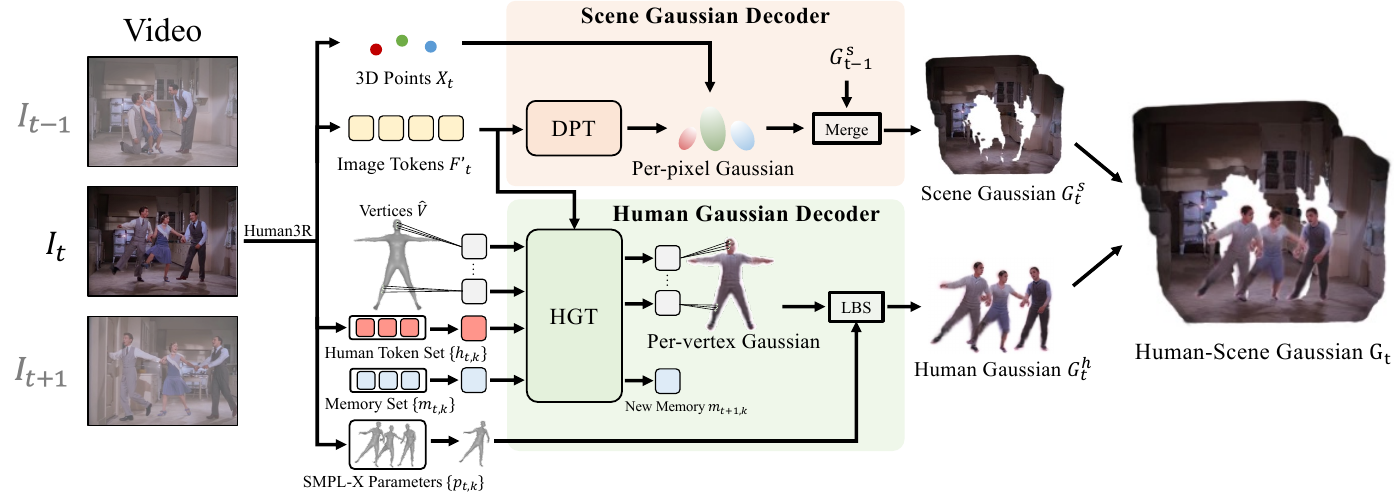}
  \caption{
  \textbf{Overview of the proposed framework.}
  GUSH3R reconstructs a dynamic human-scene representation from a monocular video using two newly introduced branches: the Scene Gaussian Decoder and the Human Gaussian Decoder.
  Each frame is processed by the foundation model Human3R~\cite{chen2025human3r} to extract human tokens $\bm{H}_t'$ and image tokens $\bm{F}_t'$ along with scene point clouds $\bm{X}_t$ and human mesh vertices $\bm{V}_t$.
  The Scene Gaussian Decoder takes the point clouds $\bm{X}_t$ as a geometric prior and predicts scene Gaussians $\bm{G}_t^{s}$ from the image token using a Dense Prediction Transformer (DPT)~\cite{dpt}.
  The Human Gaussian Decoder takes the human meshes $\bm{V}_t$ as a geometric prior and predicts human Gaussians $\bm{G}_t^{h}$ from the image and human tokens using the Human Gaussian Transformer (HGT).
  The predicted human and scene Gaussians are then merged in the same metric space as the final human-scene representation $\bm{G}_t$ and jointly rendered to produce the final image.
}
  \label{fig:architecture}
\end{figure*}
\subsection{Human-Scene Reconstruction}
Human-scene reconstruction aims to jointly recover the 3D scene geometry, human motion, and camera poses from visual observations.
Early approaches formulate this problem as a global optimization over multiple elements, including camera poses, scene geometry~\cite{dust3r,mast3rsfm,li2025megasam}, and SMPL/SMPL-X~\cite{SMPL, SMPL-X} mesh parameters~\cite{reconstructingpeopleplaces}.
Several works further adopt renderable representations such as NeRF or 3DGS~\cite{xue2024hsr,kocabas2024hugs,zhang2025odhsr,guo2023vid2avatar}, achieving high visual fidelity but requiring costly test-time optimization.

To improve efficiency, recent works explore feed-forward alternatives~\cite{rojas2025hamst3r, liu2026josh3r}.
Recent methods such as Human3R~\cite{chen2025human3r} and UniSH~\cite{unish} jointly recover static scene geometry and structured human representations in a single feed-forward pass.
While these methods provide strong geometric priors, their outputs remain heterogeneous: scenes and humans are represented as point clouds and parametric meshes, respectively.
These outputs require separate appearance modeling or representation conversion before they can be used as a complete representation for photorealistic rendering.
In contrast, our method represents both static scenes and dynamic humans as 3D Gaussians, yielding a unified and directly renderable representation suitable for downstream visual applications such as free-viewpoint rendering and immersive content creation.
\section{Method}
\subsection{Overview}
Given a monocular video $\{\bm{I}_t\}$, where $\bm{I}_t$ denotes the input frame at time $t$, our goal is to reconstruct a dynamic human-scene representation with accurate geometry and photorealistic appearance.
We decompose the entire scene $\bm{G}_{t}$ into a static scene and dynamic humans, represented as a set of scene Gaussians $\bm{G}^{\mathrm{s}}_{t}$ aggregated across time until frame $t$ and time-dependent human Gaussians $\{\bm{G}^{\mathrm{h}}_{t,k}\}$ for each frame $t$, where $k$ indexes individual humans.

We build upon a human-scene foundation model~\cite{chen2025human3r} that provides geometric priors and structured token representations.
Our key idea is to leverage these priors while disentangling appearance modeling for the static scene and dynamic humans.
Specifically, we introduce a Scene Gaussian Decoder and a Human Gaussian Decoder, detailed in Sec.~\ref{subsec:scene_appearance} and Sec.~\ref{subsec:human_appearance}, respectively.
Both decoders use Human3R geometry and tokens, enabling human and scene Gaussians to be predicted in a shared coordinate frame rather than aligned after separate reconstruction.
These designs allow us to reconstruct the full scene in a feed-forward manner from monocular input. 
An overview is shown in Fig.~\ref{fig:architecture}.

\subsection{Preliminaries}
\noindent\textbf{Human-Scene Foundation Model.}
We build upon a unified feed-forward model for human-scene reconstruction, referred to as Human3R~\cite{chen2025human3r}.
Given a monocular video $\{\bm{I}_t\}$, the model jointly estimates camera pose $\bm{T}_t$, scene point maps $\bm{X}_t$, and a set of human meshes as vertices $\{\bm{V}_{t, k}\}$ where $t$ indexes time and $k$ indexes individual humans.

Scene geometry is represented as point clouds, and human bodies are represented as parametric SMPL-X~\cite{SMPL-X} meshes.
Given the local human pose (relative axis-angle rotations) $\bm{\theta} \in \mathbb{R}^{52 \times 3}$, shape $\bm{\beta} \in \mathbb{R}^{10}$, facial expression $\bm{\alpha} \in \mathbb{R}^{10}$, and global human root transformation $\bm{P} = [\bm{R} | \bm{t}] \in \mathrm{SE}(3)$ parameterized by global orientation $\bm{R} \in \mathrm{SO}(3)$ and translation $\bm{t} \in \mathbb{R}^3$, the model outputs an expressive 3D human mesh with 10,475 vertices $\bm{V}_{t, k}$:
\begin{equation}
    \bm{V}_{t, k} =  \mathrm{SMPL\text{-}X}(\bm{\theta}, \bm{\beta}, \bm{\alpha}, \bm{P}).
\end{equation}

The backbone~\cite{oquab2024dinov2, mhmr} follows a recurrent formulation.
Each frame $\bm{I}_t$ is encoded into image tokens $\bm{F}_t$.
Then the model obtains human detections using a detector~\cite{mhmr} with a confidence threshold, which produces a set of 2D locations $\bm{u}_{t,k}$ for each frame.
Each human token $\bm{h}_{t,k}$ is obtained by sampling $\bm{F}_t$ at the detected human location $\bm{u}_{t,k}$.
All human tokens at frame $t$ are concatenated as $\bm{H}_t = \{\bm{h}_{t, k} \}$.
At the same time, the model maintains a persistent state $\bm{S}_t$ and a learnable parameter $\bm{z}_t$ corresponding to camera state, jointly updating all tokens through a decoder:
\begin{equation}
    [\bm{F}_t', \bm{z}_t', \bm{H}_t'], \bm{S}_t = \mathrm{Decoder}([\bm{F}_t, \bm{z}_t, \bm{H}_t], \bm{S}_{t-1}).
\end{equation}

From the updated image, camera, and human tokens, Human3R predicts scene point maps $\bm{X}_t$, camera poses $\bm{T}_t$, and human parameters $\{\bm{p}_{t, k}\}$ for all detected humans.
This design enables disentangled representations of the scene and humans via image and human tokens, respectively, while jointly reasoning through a shared recurrent state.
In our method, we leverage the reconstructed point clouds $\bm{X}$ and human mesh vertices $\bm{V}$ as geometric priors and further utilize both image tokens $\bm{F}$ and human tokens $\bm{H}$ to extract appearance features for downstream modeling.

\noindent\textbf{3D Gaussian Splatting.}
To achieve photorealistic rendering of both scene and humans, we adopt 3D Gaussian Splatting (3DGS)~\cite{kerbl3Dgaussians}, which represents a 3D scene as a set of anisotropic Gaussian primitives. 
Each Gaussian $\bm{g}$ models a local volumetric element and is parameterized as:
$\bm{\mu}_g, \bm{\alpha}_g, \bm{q}_g, \bm{s}_g, \bm{c}_g$
where $\bm{\mu}_g \in \mathbb{R}^3$ denotes the 3D position, $\bm{\alpha}_g \in \mathbb{R}$ represents opacity, $\bm{q}_g \in \mathbb{R}^4$ encodes rotation as a quaternion, and $\bm{s}_g \in \mathbb{R}^3$ defines anisotropic scaling. 
The appearance is represented by spherical-harmonic coefficients $\bm{c}_g \in \mathbb{R}^{3(k+1)^2}$, with $k=4$ for scene and $k=0$ for humans.

\subsection{3D Scene Reconstruction}
\label{subsec:scene_appearance}

\noindent\textbf{Scene Gaussian Decoder.}
We represent the static scene by aggregating per-frame Gaussians $\bm{g}^s_{t,i}$, each corresponding to pixel $i$ in frame $t$, into a unified static Gaussian set $\bm{G}_t^s$.
Given the reconstructed scene point maps $\bm{X}_t=\{\bm{x}_{t,i}\}$ obtained from the foundation model, we initialize each Gaussian center $\bm{\mu}_{t,i}$ directly with the corresponding 3D point $\bm{x}_{t,i}$, where $\bm{\mu}_{t, i}$ denotes the center of the Gaussian corresponding to pixel $i$ at time $t$.

Next, we decode the Human3R image tokens $\bm{F}_t'$ into per-pixel features $\bm{C}^D_{t, i}$ using a DPT-style decoder~\cite{dpt}. 
To incorporate direct appearance cues, we extract image features $\bm{C}^I_{t,i}$ from $\bm{I}_t$ using a single 7×7 convolution followed by ReLU.
We add these features to the processed DPT features and predict Gaussian parameters using a convolutional head $F_G$.
\begin{equation}
    \bm{\alpha}_{t, i}, \bm{q}_{t, i}, \bm{s}_{t, i}, \bm{c}_{t, i} = F_G(\bm{C}^D_{t, i}, \bm{C}^I_{t, i}).
\end{equation}
These parameters define a Gaussian primitive:
\begin{equation}
    \bm{g}^s_{t,i} = (\bm{\mu}_{t,i}, \bm{q}_{t,i}, \bm{s}_{t,i}, \bm{\alpha}_{t,i}, \bm{c}_{t,i}).
\end{equation}

\noindent\textbf{Filtering and Voxelization.}
At each timestep $t$, newly predicted Gaussians $\hat{\bm{G}}_t^s= \{\bm{g}^s_{t,i}\}_{i}$ are filtered using point confidence scores~\cite{cut3r} and dilated human-mask scores~\cite{mhmr}, excluding pixels whose point confidence falls below a threshold or whose human-mask scores exceed a threshold.
In addition, to handle long video sequences while maintaining memory efficiency, we introduce a voxelization scheme for the accumulated scene Gaussians inspired by AnySplat~\cite{AnySplat}.
%
%
We discretize the estimated metric coordinate system using a fixed voxel size to aggregate scene Gaussians consistently across frames.
At each timestep, newly predicted Gaussians are merged with the existing set through voxelization:
\begin{equation}
\bm{G}_t^s \leftarrow \mathrm{Voxelize}(\bm{G}_{t-1}^s, \tilde{\bm{G}}^s_t),
\end{equation}
where $\bm{G}_t^s$ denotes the accumulated scene Gaussian set up to timestep $t$, and $\tilde{\bm{G}}_t^s$ denotes the filtered per-frame Gaussians newly predicted at timestep $t$.
Within each voxel, we retain the Gaussian center corresponding to the highest confidence, while the remaining parameters are aggregated using confidence-weighted averaging. 
This filtering and voxelization strategy preserves high-quality geometry while reducing redundant scene Gaussians in repeatedly observed regions. Human Gaussians, however, are accumulated across frames, so the memory required to store them grows with sequence length (see Appendix~\ref{subsec: streaming_scalability}).

\subsection{Dynamic Human Reconstruction}
\label{subsec:human_appearance}

\noindent\textbf{Human Gaussian Decoder.}
For the $k$-th human at timestamp $t$, we use the SMPL-X~\cite{SMPL-X} mesh vertices $\bm{V}_{t, k}$ as geometric anchors for human Gaussians.
Given these vertex anchors, the Human Gaussian Decoder transfers image appearance features into the canonical body space and predicts the corresponding Gaussian attributes.
This canonicalization allows appearance cues observed under different poses and viewpoints to be associated with consistent body locations.

We define four types of tokens to capture different aspects of the human representation of the $k$-th human at timestep $t$:
\begin{itemize}
    \item \textbf{Human tokens} $\bm{h}_{t,k}$ --- representing the person-level context from Human3R (\textit{which human to attend to})
    \item \textbf{Vertex tokens} $\bm{\hat{V}}_{t,k}$ --- representing 3D SMPL-X vertices in the canonical A-pose with positional encoding (\textit{which body part it corresponds to})
    \item \textbf{Image tokens} $\bm{F}_t'$ --- extracted from the input image (\textit{what the human looks like})
    \item \textbf{Memory tokens} $\bm{m}_{t,k}$ --- storing accumulated appearance features for each person over time (\textit{what the human looked like before})
\end{itemize}
Together, these tokens provide complementary cues for person-level context, canonical body location, current-frame appearance, and temporal appearance history.

The Human Gaussian Transformer (HGT) applies self-attention jointly to human, vertex, and memory tokens, allowing information exchange among them, and cross-attention to image tokens as keys and values.
\begin{equation}
\begin{gathered}
\bm{Q}_{t,k}  = [\bm{h}_{t,k},\bm{m}_{t,k},\bm{\hat{V}}_{t,k}],\\
[\bm{m}_{t,k}',\bm{\hat{V}}_{t,k}'] = \mathrm{HGT}(Q=\bm{Q}_{t,k},K=\bm{F}_t',V=\bm{F}_t').
\end{gathered}
\end{equation}
We denote the vertex and memory features extracted from the transformer output as $\bm{\hat{V}}'_{t,k}$ and $\bm{m}'_{t,k}$, respectively.
The vertex features are fed into an MLP $F_H$ to predict Gaussian parameters for each vertex:
\begin{equation}
(\bm{\alpha}_{t,k,v}, \bm{q}_{t,k,v}, \bm{s}_{t,k,v}, \bm{c}_{t,k,v})
= F_H(\bm{\hat{V}}'_{t,k,v}),
\end{equation}
where $v$ indexes the vertices of the canonical SMPL-X mesh.
The Gaussians are defined in the canonical A-pose space and are transformed to the posed space via linear blend skinning (LBS) using SMPL-X parameters.

\noindent\textbf{Appearance Memory.}
To maintain a consistent appearance over time, we assign memory tokens $\bm{m}_{t,k}$ to each tracked person.
The memory tokens implicitly carry appearance information accumulated from previous frames and are used as additional query tokens in the Human Gaussian Decoder.
During inference, detections are associated with existing tracks by greedily matching their 2D locations in image space.
%
For each detected person, we update its memory token by interpolating between the previous token and the HGT output:
\begin{equation}
\bm{m}_{t+1,k}
= (1-\lambda)\bm{m}_{t,k}+\lambda\bm{m}_{t,k}',
\qquad \lambda=0.5.
\end{equation}
The memory is retained when the person is not detected.

\subsection{Training Details}
\noindent\textbf{Training Setup.}
We train the Scene Gaussian Decoder and the Human Gaussian Decoder separately while keeping the foundation model Human3R~\cite{chen2025human3r} frozen.
This allows each decoder to specialize in its own representation without disrupting the shared geometric prior.
In both training stages, we use consecutive frames as input to capture temporal dependencies.

\noindent\textbf{Scene Gaussian Decoder.}
We train the Scene Gaussian Decoder using the following objective:
\begin{equation}
    \mathcal{L}_{\mathrm{scene}}
    =
    \lambda_{\mathrm{mse}}\mathcal{L}_{\mathrm{mse}}
    +
    \lambda_{\mathrm{lpips}}\mathcal{L}_{\mathrm{lpips}}
    +
    \lambda_{\mathrm{dep}}\mathcal{L}_{\mathrm{dep}}
    +
    \lambda_{\mathrm{reg}}\mathcal{L}_{\mathrm{reg}}.
\end{equation}
Here, $\mathcal{L}_{\mathrm{mse}}$ and $\mathcal{L}_{\mathrm{lpips}}$~\cite{lpips} supervise the rendered scene against the input image, while $\mathcal{L}_{\mathrm{dep}}$ supervises the rendered depth against the ground truth (GT) depth to enforce geometric consistency.
For all these loss functions, we use GT masks to supervise only background regions.
$\mathcal{L}_{\mathrm{reg}}$ regularizes Gaussian scales to avoid degenerate elongated Gaussians~\cite{gaussian-reg}:
\begin{equation}
\mathcal{L}_{\mathrm{reg}}
=
\frac{1}{N}\sum_{i}
\max\!\left(
\frac{\max(\bm{s}_i)}{\min(\bm{s}_i)} - \tau,\; 0
\right),
\end{equation}
where $N$ is the number of scene Gaussians included, $\bm{s}_i$ denotes the scale parameters of the $i$-th Gaussian, and $\tau=3$ is the maximum scale ratio
allowed without penalty.

We train the Scene Gaussian Decoder on the BEDLAM~\cite{bedlam} and DL3DV~\cite{dl3dv}, sampling the two datasets with equal probability.
Each training sample contains four frames at uniform intervals.
Both datasets provide RGB and GT depth supervision.

\noindent\textbf{Human Gaussian Decoder.}
We train the Human Gaussian Decoder using the following objective, which is defined for each person at each frame:
\begin{equation}
    \mathcal{L}_{\mathrm{human}}
    =
    \lambda_{\mathrm{mse}}\mathcal{L}_{\mathrm{mse}}
    +
    \lambda_{\mathrm{part}}\mathcal{L}_{\mathrm{part}}
    +
    \lambda_{\mathrm{sil}}\mathcal{L}_{\mathrm{sil}}
    +
    \lambda_{\mathrm{reg}}\mathcal{L}_{\mathrm{reg}}.
\end{equation}
Here, $\mathcal{L}_{\mathrm{mse}}$ and $\mathcal{L}_{\mathrm{sil}}$ supervise the rendered human appearance and silhouette using pixel-wise MSE and binary cross entropy, respectively, while $\mathcal{L}_{\mathrm{reg}}$ penalizes degenerate Gaussian shapes as in the Scene Gaussian Decoder.
We additionally use a partial LPIPS loss to improve fine-grained human appearance:
\begin{equation}
\begin{aligned}
\mathcal{L}_{\mathrm{part}}
={}&
\mathrm{LPIPS}(\hat{\bm{I}}, \bm{I}) \\
&+
\sum_{r \in \{\mathrm{upper},\,\mathrm{face}\}}
\mathrm{LPIPS}\bigl(
\mathrm{crop}_r(\hat{\bm{I}}),
\mathrm{crop}_r(\bm{I})
\bigr).
\end{aligned}
\end{equation}
Here, $\hat{\bm{I}}$ and $\bm{I}$ denote the rendered and input images, and $\mathrm{crop}_r(\cdot)$ extracts the upper-body or face region.
We train the Human Gaussian Decoder on BEDLAM~\cite{bedlam} and
MVHumanNet~\cite{Xiong2024MVHumanNet}, sampling the two datasets
with equal probability. Each training sample contains four frames
at uniform intervals.

\section{Experiments}
\label{sec: exp}
\subsection{Experimental Setups}
We evaluate our method on dynamic human-scene reconstruction, focusing on the photorealistic rendering quality for both dynamic humans and static scenes.
Since no existing feed-forward method directly addresses our setting, we compare against both an optimization-based human-scene method and decomposition-based feed-forward baselines.
For the optimization-based baseline, we use HSR~\cite{xue2024hsr}, which provides publicly available code.
For the decomposition-based baselines, we reconstruct the static background with AnySplat~\cite{AnySplat}, reconstruct dynamic humans with LHM~\cite{qiu2025LHM}, and compose them in a common coordinate frame.
We report three variants: AnySplat, AnySplat+LHM+Human3R, and AnySplat+LHM+GT Pose.
AnySplat does not explicitly model humans, while the latter two use SMPL-X poses from Human3R and GT, respectively.
We also make comparisons with general 4D reconstruction methods on novel view video synthesis.
For quantitative evaluation, we report PSNR, SSIM~\cite{ssim}, and LPIPS~\cite{lpips} to measure novel-view rendering quality, as well as reconstruction time, including I/O, scene encoding, recurrent state updates, and Gaussian parameter prediction, to evaluate efficiency.

For video-based evaluation, we additionally report optical-flow-based warping error~\cite{warp} to measure temporal consistency.
%
Further details of training and the baselines are provided in Appendix~\ref{subsec:app_training} and Appendix~\ref{subsec:app_baselines}, respectively.
Long-sequence streaming performance is analyzed in Appendix~\ref{subsec: streaming_scalability}.

\subsection{Human-Scene Reconstruction}
\subsubsection{Single-Human Scene}
\label{subsec:exp_single_nvs}
We evaluate the quality of novel view synthesis on NeuMan~\cite{NeuMan} and EMDB~\cite{emdb}, both of which contain monocular videos of dynamic single-human scenes.
We alternate input and target frames in temporal order and reconstruct our Gaussians using only the input frames.
Each target image is rendered from its GT camera parameters using Gaussians animated with the target SMPL-X parameters.
For our method, target SMPL-X parameters are estimated in a separate Human3R pass over the entire evaluation sequence, including both input and target frames. 
Target RGB is used for pose estimation but not for Gaussian appearance reconstruction.
The Human3R and GT Pose baselines use Human3R-estimated and ground-truth target SMPL-X, respectively.
\begin{figure}[t]
  \centering
  \includegraphics[width=\linewidth]{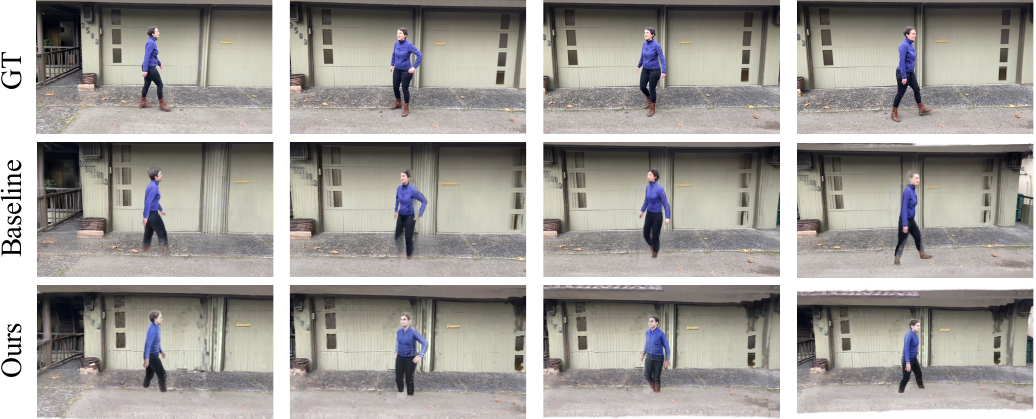}
  \caption{
    \textbf{Qualitative comparison of single-human scene reconstruction against the baseline using NeuMan~\cite{NeuMan}.}
    The baseline refers to the decomposition-based combination of AnySplat~\cite{AnySplat}, LHM~\cite{qiu2025LHM}, and Human3R~\cite{chen2025human3r}.
    Although our method reconstructs Gaussians using only past frames, it achieves comparable reconstruction quality while providing faster inference.
  }
  \label{fig:single_human_scene}
\end{figure}
As shown in Fig.~\ref{fig:single_human_scene}, decomposition-based baselines often leave visible artifacts around humans and scene boundaries, reflecting the difficulty of aligning separately reconstructed humans and scenes.
In contrast, our method produces coherent human-scene renderings while maintaining comparable visual quality for both the human body and the surrounding scene.

\begin{table}[t]
\centering
\renewcommand{\arraystretch}{1.12}
\setlength{\tabcolsep}{4pt}
\resizebox{\linewidth}{!}{%
\begin{tabular}{l ccc ccc c}
\toprule
\textbf{Method}
& \multicolumn{3}{c}{\textbf{NeuMan}}
& \multicolumn{3}{c}{\textbf{EMDB}}
& \textbf{Recon. Time} \\
\cmidrule(lr){2-4}
\cmidrule(lr){5-7}
& PSNR$\uparrow$ & SSIM$\uparrow$ & LPIPS$\downarrow$
& PSNR$\uparrow$ & SSIM$\uparrow$ & LPIPS$\downarrow$
&  (/frame)$\downarrow$  \\
\midrule

\multicolumn{8}{l}{\textit{Optimization-based method}} \\
HSR
& \textbf{20.6} & \textbf{0.58} & 0.58
& \underline{16.2} & \textbf{0.68} & 0.51
&  $>$ 1h \\

\specialrule{1.0pt}{2pt}{2pt}
\multicolumn{8}{l}{\textit{Feed-forward Baselines}} \\

AnySplat
& 15.2 & 0.33 & \underline{0.42}
& 13.2 & 0.46 & 0.51
& \textbf{0.47 s} \\

AnySplat+LHM+Human3R
& 13.9 & 0.32 & 0.46
& 14.7 & 0.48 & \underline{0.49}
& 5.5 s \\

AnySplat+LHM+GT Pose
& {\color{lightgray}14.6}
& {\color{lightgray}0.32}
& {\color{lightgray}0.43}
& {\color{lightgray}12.4}
& {\color{lightgray}0.45}
& {\color{lightgray}0.52}
& {\color{lightgray} 4.2 s} \\

\specialrule{1.0pt}{2pt}{2pt}
\textbf{Ours}
& \underline{18.2} & \underline{0.57} & \textbf{0.29}
& \textbf{18.0} & \underline{0.57} & \textbf{0.40}
& \underline{0.54 s} \\

\bottomrule
\end{tabular}%
}
\caption{
\textbf{Single-human scene novel view synthesis on NeuMan~\cite{NeuMan} and EMDB~\cite{emdb}. } Reconstruction time excludes the additional Human3R pass used to estimate target-frame human poses.}
\label{tab:single_human_nvs}
\end{table}

As shown in Table~\ref{tab:single_human_nvs}, HSR achieves higher PSNR on NeuMan and higher SSIM on both datasets, whereas our method achieves higher PSNR on EMDB and lower LPIPS on both datasets, with a substantially shorter reconstruction time.
Detailed breakdowns of the reconstruction time are in Appendix~\ref{subsec: runtime_breakdown}.
Compared with the two baselines that combine AnySplat and LHM, our method improves both rendering quality and reconstruction time, suggesting that joint human-scene reconstruction is more effective than post-hoc composition.

\subsubsection{Multi-Human Scene}
\label{subsec:exp_multi_nvs}
\begin{figure}[t]
  \centering
  \includegraphics[width=\linewidth]{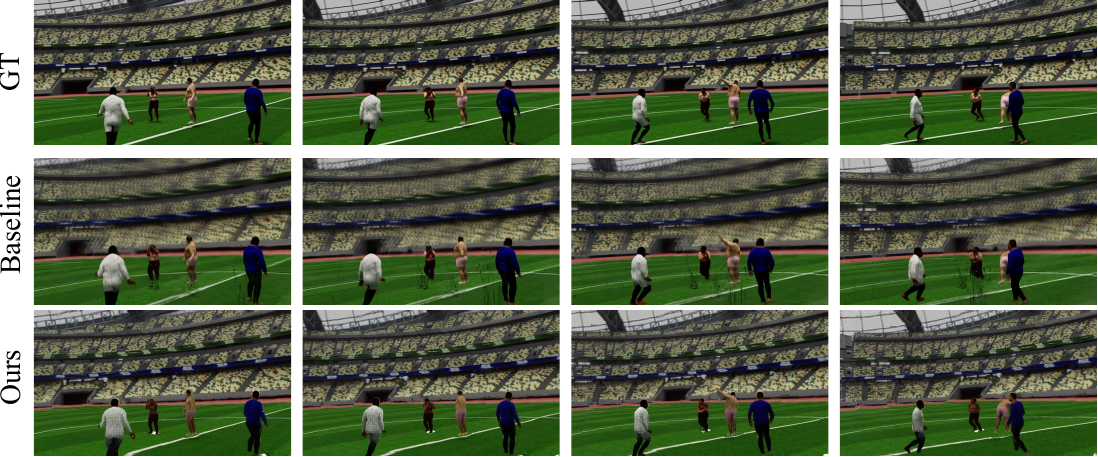}
  \caption{
    \textbf{Qualitative comparison on multi-human scene reconstruction against the baseline using BEDLAM~\cite{bedlam}.}
    The baseline approach refers to the decomposition-based combination of AnySplat~\cite{AnySplat}, LHM~\cite{qiu2025LHM}, and Human3R~\cite{chen2025human3r}.
  }
  \label{fig:multi_human_scene}
\end{figure}

We use the same evaluation protocol as for NeuMan and EMDB.
We begin with qualitative comparisons of novel view synthesis on the BEDLAM~\cite{bedlam} test split against the decomposition-based baselines.
As shown in Fig.~\ref{fig:multi_human_scene}, the decomposition-based baselines can reconstruct the background to some extent and humans in good quality, but often suffer from inaccurate human-scene alignment and visible composition artifacts.
In contrast, our method reconstructs the static environment and multiple dynamic humans within a unified 3D Gaussian representation.

We then quantify these observations about novel view synthesis on the BEDLAM test split.
\begin{table}[t]
\centering
\renewcommand{\arraystretch}{1.15}
\resizebox{\linewidth}{!}{%
\begin{tabular}{l ccc ccc ccc}
\toprule
\textbf{Method}
& \multicolumn{3}{c}{\textbf{Human-Scene}}
& \multicolumn{3}{c}{\textbf{Scene}}
& \multicolumn{3}{c}{\textbf{Human}} \\
\cmidrule(lr){2-4}
\cmidrule(lr){5-7}
\cmidrule(lr){8-10}
& PSNR$\uparrow$ & SSIM$\uparrow$ & LPIPS$\downarrow$
& PSNR$\uparrow$ & SSIM$\uparrow$ & LPIPS$\downarrow$
& PSNR$\uparrow$ & SSIM$\uparrow$ & LPIPS$\downarrow$ \\
\midrule

AnySplat
& 15.9 & 0.43 & 0.42
& 16.2 & 0.50 & 0.37
& 14.3 & 0.87 & 0.14 \\

AnySplat+LHM+Human3R
& 14.5 & 0.31 & 0.47
& 14.5 & 0.38 & 0.46
& \textbf{14.7} & \textbf{0.88} & \textbf{0.11} \\

AnySplat+LHM+GT Pose
& {\color{lightgray}15.1} & {\color{lightgray}0.35} & {\color{lightgray}0.43}
& {\color{lightgray}15.0} & {\color{lightgray}0.40} & {\color{lightgray}0.42}
& {\color{lightgray}16.9} & {\color{lightgray}0.90} & {\color{lightgray}0.08} \\

\textbf{Ours}
& \textbf{17.0} & \textbf{0.53} & \textbf{0.34}
& \textbf{17.5} & \textbf{0.59} & \textbf{0.30}
& 14.2 & \textbf{0.88} & \textbf{0.11} \\

\bottomrule
\end{tabular}%
}
\caption{
\textbf{Multi-human scene novel view synthesis on BEDLAM~\cite{bedlam}.}
We evaluate the full image (\textbf{Human-Scene}), background regions (\textbf{Scene}), and human regions (\textbf{Human}) against the decomposition-based baselines.
}
\label{tab:multi_human_nvs}
\end{table}
As shown in Table~\ref{tab:multi_human_nvs}, our method achieves the best performance on the full human-scene evaluation, demonstrating the benefit of avoiding post-hoc composition.
On human regions, our method matches the SSIM and LPIPS of AnySplat+LHM+Human3R but achieves lower PSNR (14.2 vs. 14.7), despite improving full-scene quality.

\begin{figure*}[t]
  \centering
  \includegraphics[width=\linewidth]{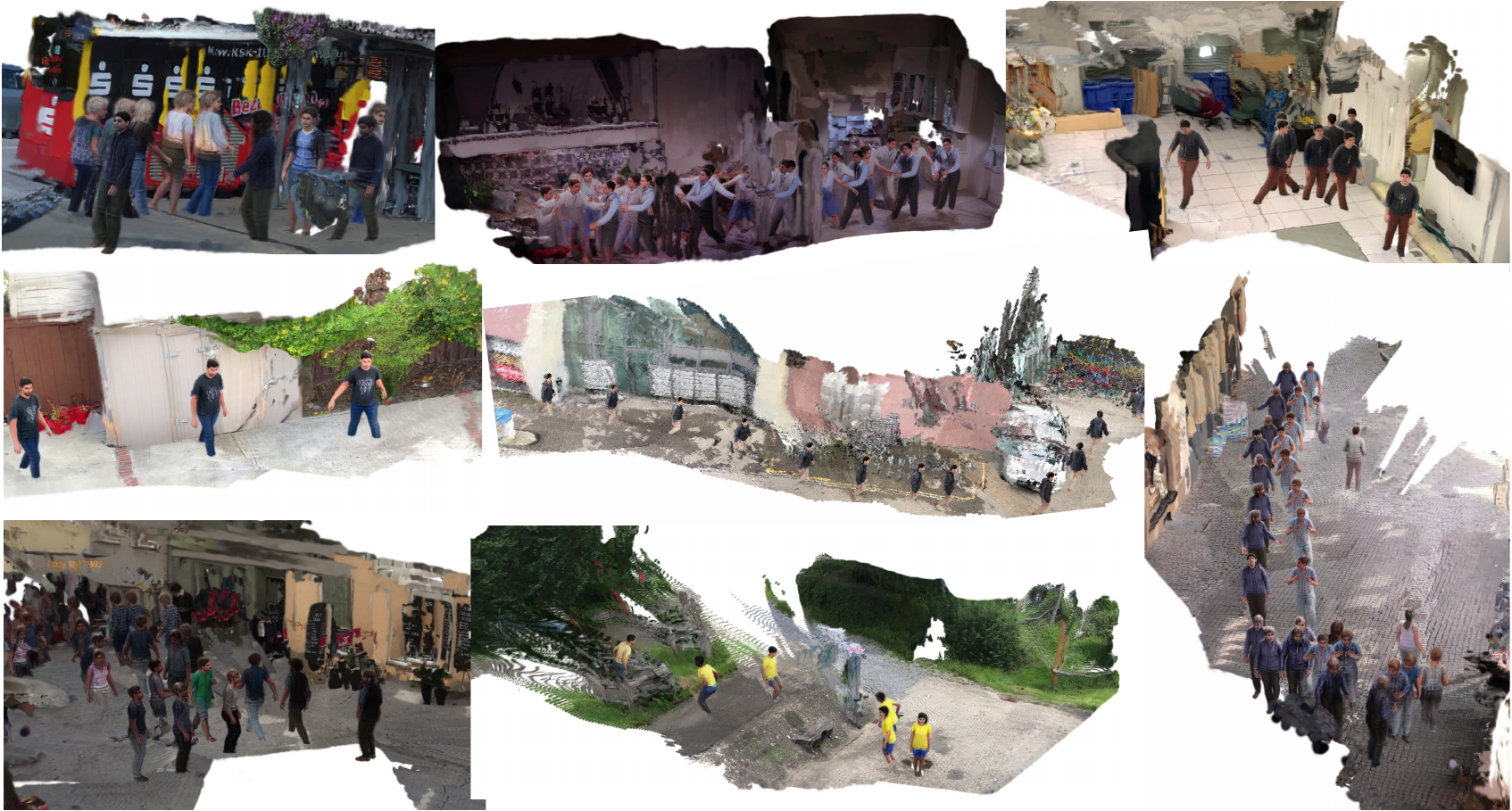}
  \caption{
    \textbf{Qualitative 4D human-scene reconstruction results.}
    GUSH3R produces coherent dynamic human-scene reconstructions across diverse scenarios~\cite{emdb, NeuMan, 3dpw} with different numbers of people, body poses, and camera viewpoints from a monocular video.
  }
  \label{fig:multi_human_gallery}
\end{figure*}

Finally, Fig.~\ref{fig:multi_human_gallery} shows additional qualitative results beyond the BEDLAM test split.

\subsection{Novel View Video Synthesis}
\label{subsec:nvvs}
We evaluate novel view video synthesis from monocular videos on MultiEgo~\cite{multiego}.
Each method reconstructs from a monocular input video and renders synchronized views from a held-out camera. 
Our method estimates human poses from the input view without accessing target RGB.

We compare against two classes of methods.
First, we evaluate recent feed-forward static reconstruction methods, including AnySplat~\cite{AnySplat}, YoNoSplat~\cite{yonosplat}, and Depth Anything 3 (DA3)~\cite{depthanything3}.
Although these methods do not explicitly model scene dynamics, they provide strong geometry and appearance priors and therefore serve as informative baselines for quantifying the importance of temporal modeling.

Second, we compare against dynamic reconstruction methods.
Shape of Motion~\cite{som2024} performs per-scene optimization from a monocular video, whereas MoVieS~\cite{movies}, 4DGT~\cite{xu20254dgt}, and StreamSplat~\cite{wu2025streamsplat} reconstruct dynamic scenes in a feed-forward manner.
Among these methods, Shape of Motion, MoVieS, and 4DGT require posed input videos, while the remaining methods can operate without GT input camera parameters.
StreamSplat and our method additionally operate in a streaming manner, processing input frames sequentially rather than jointly reconstructing the full input sequence.
For methods that estimate input camera poses, we align the reconstructed
coordinate system with the GT coordinate system using a single
global similarity transformation estimated from the input cameras.

\begin{table}[t]
\centering
\renewcommand{\arraystretch}{1.15}
\setlength{\tabcolsep}{4.5pt}
\resizebox{\linewidth}{!}{%
\begin{tabular}{l ccc ccc c c}
\toprule
\textbf{Method}
& \textbf{Feed-Forward}
& \textbf{Pose-Free}
& \textbf{Streaming}
& \textbf{PSNR}$\uparrow$
& \textbf{SSIM}$\uparrow$
& \textbf{LPIPS}$\downarrow$
& \textbf{Warp Err.}$\downarrow$
& \textbf{Time / 8 frames (s)}$\downarrow$ \\

\midrule

\multicolumn{9}{l}{\textit{Static novel view synthesis}} \\

AnySplat~\cite{AnySplat}
& {\color{Green}\ding{51}}
& {\color{Green}\ding{51}}
& {\color{red}\ding{55}}
& 11.6
& 0.62
& 0.47
& -- & \underline{1.15} \\

YoNoSplat~\cite{yonosplat}
& {\color{Green}\ding{51}}
& {\color{Green}\ding{51}}
& {\color{red}\ding{55}}
& 11.8 & 0.66 & 0.48 & -- & \textbf{1.01} \\

DA3~\cite{depthanything3}
& {\color{Green}\ding{51}}
& {\color{Green}\ding{51}}
& {\color{red}\ding{55}}
& \underline{13.9}
& \underline{0.68}
& \underline{0.45}
& -- & 2.26 \\

\midrule

\multicolumn{9}{l}{\textit{Dynamic novel view video synthesis}} \\

Shape of Motion~\cite{som2024}
& {\color{red}\ding{55}}
& {\color{red}\ding{55}}
& {\color{red}\ding{55}}
& 12.2 & 0.61 & 0.47 & 0.0030 & 274.3 \\

MoVieS~\cite{movies}
& {\color{Green}\ding{51}}
& {\color{red}\ding{55}}
& {\color{red}\ding{55}}
& 12.9
& 0.61
& \underline{0.45}
& 0.0025 & 2.02 \\

4DGT~\cite{xu20254dgt}
& {\color{Green}\ding{51}}
& {\color{red}\ding{55}}
& {\color{red}\ding{55}}
& 9.1
& 0.46
& 0.50
& \underline{0.0022} & 1.19 \\

StreamSplat~\cite{wu2025streamsplat}
& {\color{Green}\ding{51}}
& {\color{Green}\ding{51}}
& {\color{Green}\ding{51}}
& 13.2 & 0.49 & 0.52 & 0.0099 & 5.91 \\

\midrule

\textbf{Ours}
& {\color{Green}\ding{51}}
& {\color{Green}\ding{51}}
& {\color{Green}\ding{51}}
& \textbf{14.2}
& \textbf{0.70}
& \textbf{0.38}
& \textbf{0.0015}
& 5.50 \\

\bottomrule
\end{tabular}%
}
\caption{
\textbf{Novel view video synthesis comparison on MultiEgo~\cite{multiego}.}
We compare static and dynamic reconstruction methods using publicly available implementations.
``Pose-Free'' indicates whether camera poses are not required as input, and ``Streaming'' indicates whether frames can be processed sequentially without the full input sequence.
All methods are evaluated by rendering the same held-out GT target camera trajectory.
Reconstruction time is measured for eight input frames on a single NVIDIA A100 GPU.
}
\label{tab:nvvs}
\end{table}

As shown in Table~\ref{tab:nvvs}, our method achieves the best overall rendering quality among all evaluated baselines. Compared with the strongest static baseline, DA3, our method improves PSNR from 13.9 to 14.2 and SSIM from 0.68 to 0.70, while reducing LPIPS from 0.45 to 0.38. This suggests that explicitly modeling scene dynamics is important for videos containing moving humans, where a single static representation cannot fully explain temporally varying observations.

Our method outperforms dynamic feed-forward baselines while remaining pose-free, unlike MoVieS and 4DGT, which require input camera poses.
Furthermore, our method obtains the lowest warping error of 0.0015, indicating strong temporal consistency.
In terms of efficiency, several non-streaming feed-forward methods achieve lower reconstruction times by processing all eight input frames jointly.
In contrast, our method processes frames sequentially in a streaming manner.
Under this comparable streaming setting, our method is slightly faster than StreamSplat (5.50 s vs. 5.91 s for eight frames), while substantially improving rendering quality and temporal consistency.
Overall, these results demonstrate a favorable balance between rendering quality, temporal consistency, pose-free reconstruction, and efficiency.

\subsection{Model Analysis and Ablation Studies}
\label{subsec:ablations}

\subsubsection{Component Ablations}
We analyze the components of the Scene and Human Gaussian Decoders under various input settings on NeuMan~\cite{NeuMan}, as shown in Table~\ref{tab:ablation_combined}.

\begin{table}[t]
\centering
\renewcommand{\arraystretch}{1.15}
\resizebox{\linewidth}{!}{%
\begin{tabular}{l ccc ccc ccc}
\toprule
\textbf{Setting / Variant}
& \multicolumn{3}{c}{\textbf{4-view}}
& \multicolumn{3}{c}{\textbf{8-view}}
& \multicolumn{3}{c}{\textbf{16-view}} \\
\cmidrule(lr){2-4}
\cmidrule(lr){5-7}
\cmidrule(lr){8-10}
& PSNR$\uparrow$ & SSIM$\uparrow$ & LPIPS$\downarrow$
& PSNR$\uparrow$ & SSIM$\uparrow$ & LPIPS$\downarrow$
& PSNR$\uparrow$ & SSIM$\uparrow$ & LPIPS$\downarrow$ \\
\midrule

\multicolumn{10}{l}{\textit{Scene}} \\

\textbf{Full model}
& \textbf{19.7} & \textbf{0.60} & \textbf{0.26}
& \textbf{17.8} & \textbf{0.49} & \textbf{0.37}
& \textbf{17.4} & \textbf{0.47} & \textbf{0.39} \\

{\color{red}w/o} depth loss
& 19.3 & 0.58 & 0.29
& 17.6 & 0.48 & 0.40
& 17.1 & 0.46 & 0.43 \\

{\color{red}w/o} DL3DV~\cite{dl3dv}
& 19.5 & 0.59 & 0.28
& 17.7 & 0.48 & \textbf{0.37}
& 17.3 & \textbf{0.47} & \textbf{0.39} \\

\midrule

\multicolumn{10}{l}{\textit{Human}} \\
\textbf{Full model}
& \textbf{12.6} & \textbf{0.76} & \textbf{0.18}
& \textbf{13.7} & \textbf{0.81} & \textbf{0.15}
& \textbf{13.2} & \textbf{0.78} & \textbf{0.17} \\

{\color{red}w/o} MVHumanNet~\cite{Xiong2024MVHumanNet}
& 12.4 & 0.75 & 0.20
& 13.4 & 0.80 & 0.16
& 13.0 & 0.77 & 0.19 \\

{\color{red}w/o} memory tokens
& 12.5 & \textbf{0.76} & \textbf{0.18}
& 13.6 & 0.80 & 0.16
& \textbf{13.2} & \textbf{0.78} & \textbf{0.17} \\

{\color{red}w/o} cross-attention
& 11.0 & 0.73 & 0.22
& 12.4 & 0.79 & 0.20
& 11.8 & 0.76 & 0.21 \\

\bottomrule
\end{tabular}%
}
\caption{
\textbf{Ablation study on Scene Gaussian Decoder and Human Gaussian Decoder.}
We report performance under different view settings on NeuMan~\cite{NeuMan}.
}
\label{tab:ablation_combined}
\end{table}

\noindent\textbf{Scene Gaussian Decoder.}
GT depth supervision consistently improves PSNR, SSIM, and LPIPS, indicating that explicit geometric guidance helps regularize Gaussian placement and improves scene reconstruction.
In contrast, adding DL3DV~\cite{dl3dv} provides only marginal gains, suggesting that stable geometry is more critical than additional appearance diversity in this setting.
Further analyses of depth supervision and static-background reconstruction are provided in Appendix~\ref{subsec:exp_depth} and Appendix~\ref{subsec:exp_static}, respectively.

\noindent\textbf{Human Gaussian Decoder.}
Removing cross-attention leads to the largest degradation, highlighting its importance for transferring image appearance to canonical human Gaussians.
MVHumanNet~\cite{Xiong2024MVHumanNet} and memory tokens further improve rendering quality.

\subsubsection{Cross-Attention Analysis}
\begin{figure}[t]
    \centering
    \includegraphics[width=\linewidth]
    {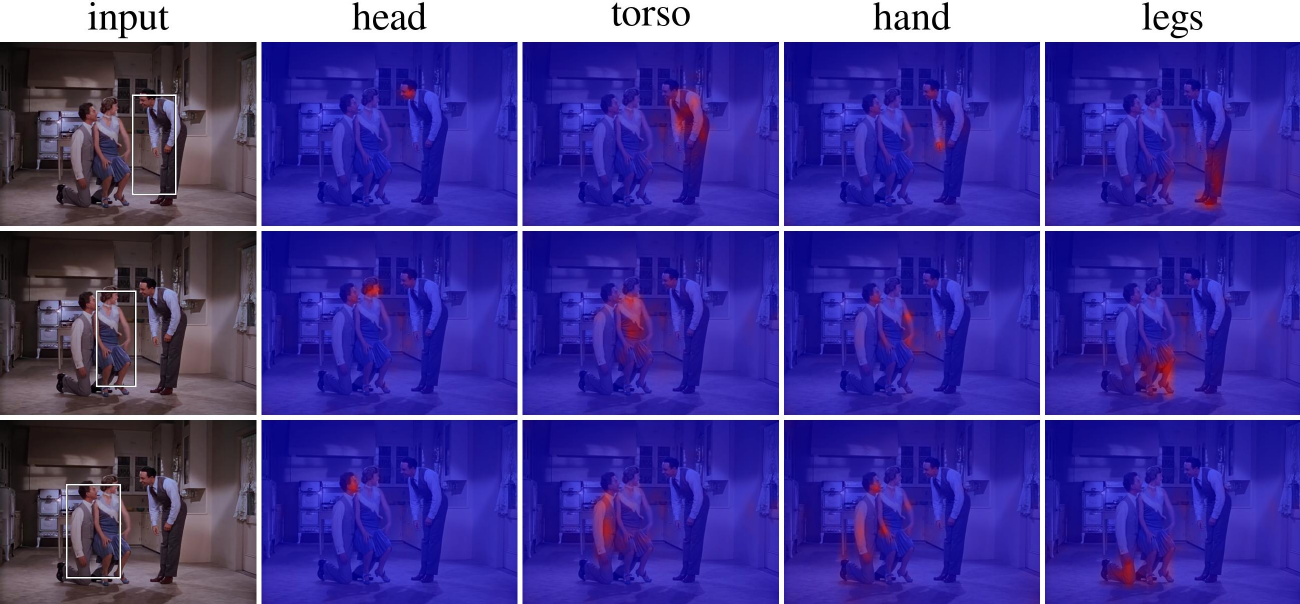}
    \caption{
    \textbf{Person-specific cross-attention visualization.}
    Each row represents one person, and the columns show the target bounding box and attention maps for the head, torso, hand, and legs.
    Warmer colors indicate higher attention to the corresponding image patches.
    }
    \label{fig:cross_attention}
\end{figure}

Fig.~\ref{fig:cross_attention} visualizes cross-attention from SMPL-X-conditioned queries to image patches. 
The attention weights are averaged over transformer layers, attention heads, and queries belonging to each anatomical region.
These results indicate that the decoder learns person-specific and anatomically structured correspondences, although some ambiguity remains between nearby or overlapping subjects.
\section{Conclusion}
We present GUSH3R, a feed-forward method for streaming dynamic human-scene reconstruction from monocular video. 
By lifting pretrained geometric and human priors into scene and human Gaussians, it enables photorealistic rendering without per-scene optimization.
Experiments show improved efficiency over optimization-based baselines and strong novel view video synthesis under a pose-free, streaming setting.
Limitations and future directions are discussed in Appendix~\ref{sec:limitation}.
\clearpage
\section*{Acknowledgments}
This work was partially supported by JST ASPIRE Program, Japan, Grant Number
JPMJAP2303; JST ACT-X (JPMJAX25C5); JST SPRING, Grant Number JPMJSP2108; and JSPS
KAKENHI Grant Number 26K21245.

{
    \small
    \bibliographystyle{ieeenat_fullname}
    \bibliography{main}

@article{AR-splatloc,
  title={{SplatLoc}: {3D} {Gaussian} Splatting-based Visual Localization for Augmented Reality},
  author={Zhai, Hongjia and Zhang, Xiyu and Zhao, Boming and Li, Hai and He, Yijia and Cui, Zhaopeng and Bao, Hujun and Zhang, Guofeng},
  journal={TVCG},
  volume={31},
  number={5},
  pages={3591-3601},
  year={2025},
}

@article{SMPL,
  author = {Loper, Matthew and Mahmood, Naureen and Romero, Javier and Pons-Moll, Gerard and Black, Michael J.},
  title = {{SMPL}: A Skinned Multi-Person Linear Model},
  journal = {TOG},
  volume={34},
  number={6},
  pages={248:1--248:16},
  articleno={248},
  year = {2015}
}

@inproceedings{SMPL-X,
  title = {Expressive Body Capture: {3D} Hands, Face, and Body from a Single Image},
  author = {Pavlakos, Georgios and Choutas, Vasileios and Ghorbani, Nima and Bolkart, Timo and Osman, Ahmed A. A. and Tzionas, Dimitrios and Black, Michael J.},
  booktitle = {CVPR},
  pages = {10975--10985},
  year = {2019}
}

@inproceedings{rome,
  title={Building {Rome} in a day},
  author={Agarwal, Sameer and Snavely, Noah and Simon, Ian and Seitz, Steven M. and Szeliski, Richard},
  booktitle={ICCV},
  pages={72-79},
  year={2009},
}

@inproceedings{romecloudlessday,
  title={Building {Rome} on a cloudless day},
  author = {Frahm, Jan-Michael and Fite-Georgel, Pierre and Gallup, David and Johnson, Tim and Raguram, Rahul and Wu, Changchang and Jen, Yi-Hung and Dunn, Enrique and Clipp, Brian and Lazebnik, Svetlana and Pollefeys, Marc},
  booktitle={ECCV},
  series = {Lecture Notes in Computer Science},
  volume = {6314},
  pages = {368--381},
  year={2010},
}

@inproceedings{robustsfm,
  title={Robust incremental structure-from-motion with hybrid features},
  author={Liu, Shaohui and Gao, Yidan and Zhang, Tianyi and Pautrat, R{\'e}mi and Sch{\"o}nberger, Johannes L and Larsson, Viktor and Pollefeys, Marc},
  booktitle={ECCV},
series = {Lecture Notes in Computer Science},
volume = {15094},
pages = {249--269},
  year={2024},
}

@inproceedings{colmap,
  title={Structure-from-motion revisited},
  author={Sch{\"o}nberger, Johannes L and Frahm, Jan-Michael},
  booktitle={CVPR},
  pages = {4104--4113},
  year={2016}
}

@inproceedings{galliani2015massive-mvs,
  title={Massively parallel multiview stereopsis by surface normal diffusion},
  author={Galliani, Silvano and Lasinger, Katrin and Schindler, Konrad},
  booktitle={ICCV},
  pages = {873--881},
  year={2015}
}

@inproceedings{pixelwise,
  title={Pixelwise view selection for unstructured multi-view stereo},
  author={Sch{\"o}nberger, Johannes L and Zheng, Enliang and Frahm, Jan-Michael and Pollefeys, Marc},
  booktitle={ECCV},
series = {Lecture Notes in Computer Science},
volume = {9907},
pages = {501--518},
  year={2016},
}

@inproceedings{mildenhall2020nerf,
  title={{NeRF}: Representing Scenes as Neural Radiance Fields for View Synthesis},
  author={Mildenhall, Ben and Srinivasan, Pratul P and Tancik, Matthew and Barron, Jonathan T and Ramamoorthi, Ravi and Ng, Ren},
  booktitle={ECCV},
series = {Lecture Notes in Computer Science},
volume = {12346},
pages = {405--421},
  year={2020},
}

@article{kerbl3Dgaussians,
  title={{3D} {Gaussian} splatting for real-time radiance field rendering},
  author={Kerbl, Bernhard and Kopanas, Georgios and Leimk{\"u}hler, Thomas and Drettakis, George},
  journal={TOG},
  volume = {42},
  number = {4},
  artcileno = {139},
  pages={139:1--139:14},
  year={2023}
}

@inproceedings{dust3r,
  title={{DUSt3R}: Geometric {3D} Vision Made Easy},
  author={Wang, Shuzhe and Leroy, Vincent and Cabon, Yohann and Chidlovskii, Boris and Revaud, Jerome},
  booktitle={CVPR},
  pages = {20697--20709},
  year={2024}
}

@inproceedings{mast3r,
  title={Grounding image matching in {3D} with {MASt3R}},
  author={Leroy, Vincent and Cabon, Yohann and Revaud, Jerome},
  booktitle={ECCV},
  series = {Lecture Notes in Computer Science},
  volume = {15130},
  pages = {71--91},
  year={2024},
}

@inproceedings{mast3rsfm,
  title={{MASt3R-SFM: A} fully-integrated solution for unconstrained structure-from-motion},
  author={Duisterhof, Bardienus Pieter and Zust, Lojze and Weinzaepfel, Philippe and Leroy, Vincent and Cabon, Yohann and Revaud, Jerome},
  booktitle={3DV},
  pages = {1--10},
  year={2025},
}

@inproceedings{wang2025vggt,
  title={{VGGT}: Visual Geometry Grounded Transformer},
  author={Wang, Jianyuan and Chen, Minghao and Karaev, Nikita and Vedaldi, Andrea and Rupprecht, Christian and Novotny, David},
  booktitle={CVPR},
  pages = {5294--5306},
  year={2025}
}

@inproceedings{vit,
  title={An image is worth 16x16 words: Transformers for image recognition at scale},
  author={Alexey Dosovitskiy and Lucas Beyer and Alexander Kolesnikov and Dirk Weissenborn and Xiaohua Zhai and Thomas Unterthiner and Mostafa Dehghani and Matthias Minderer and Georg Heigold and Sylvain Gelly and Jakob Uszkoreit and Neil Houlsby},
  booktitle={ICLR},
  year={2021}
}

@article{oquab2024dinov2,
title={{DINO}v2: Learning Robust Visual Features without Supervision},
author={Maxime Oquab and Timoth{\'e}e Darcet and Th{\'e}o Moutakanni and Huy V. Vo and Marc Szafraniec and Vasil Khalidov and Pierre Fernandez and Daniel Haziza and Francisco Massa and Alaaeldin El-Nouby and Mido Assran and Nicolas Ballas and Wojciech Galuba and Russell Howes and Po-Yao Huang and Shang-Wen Li and Ishan Misra and Michael Rabbat and Vasu Sharma and Gabriel Synnaeve and Hu Xu and Herve Jegou and Julien Mairal and Patrick Labatut and Armand Joulin and Piotr Bojanowski},
  journal={TMLR},
  issn={2835-8856},
  year={2024}
}

@inproceedings{4dvisofdynamicevents,
  title={{4D} visualization of dynamic events from unconstrained multi-view videos},
  author={Bansal, Aayush and Vo, Minh and Sheikh, Yaser and Ramanan, Deva and Narasimhan, Srinivasa},
  booktitle={CVPR},
  pages={5365–5374},
  year={2020}
}

@inproceedings{cao2023hexplane,
  title={{HexPlane}: A fast representation for dynamic scenes},
  author={Cao, Ang and Johnson, Justin},
  booktitle={CVPR},
  pages = {130--141},
  year={2023}
}

@inproceedings{neural3dvidefrommultiview,
  title={Neural {3D} video synthesis from multi-view video},
  author={Li, Tianye and Slavcheva, Mira and Zollhoefer, Michael and Green, Simon and Lassner, Christoph and Kim, Changil and Schmidt, Tanner and Lovegrove, Steven and Goesele, Michael and Newcombe, Richard and Lv, Zhaoyang},
  booktitle={CVPR},
  pages = {5521--5531},
  year={2022}
}

@inproceedings{pumarola2020d-dnerf,
  title={{D-NeRF}: Neural Radiance Fields for Dynamic Scenes},
  author={Pumarola, Albert and Corona, Enric and Pons-Moll, Gerard and Moreno-Noguer, Francesc},
  booktitle={CVPR},
  pages = {10318--10327},
  year={2021}
}

@InProceedings{Wu_2024_CVPR-4dgs,
  title={{4D} {Gaussian} splatting for real-time dynamic scene rendering},
  author={Wu, Guanjun and Yi, Taoran and Fang, Jiemin and Xie, Lingxi and Zhang, Xiaopeng and Wei, Wei and Liu, Wenyu and Tian, Qi and Wang, Xinggang},
  booktitle={CVPR},
  pages = {20310--20320},
  year={2024}
}

@inproceedings{bullettime,
  title={Feed-Forward Bullet-Time Reconstruction of Dynamic Scenes from Monocular Videos},
  author={Liang, Hanxue and Ren, Jiawei and Mirzaei, Ashkan and Torralba, Antonio and Liu, Ziwei and Gilitschenski, Igor and Fidler, Sanja and Oztireli, Cengiz and Ling, Huan and Gojcic, Zan and Huang, Jiahui},
  booktitle={NeurIPS},
  volume = {38},
  pages = {151718--151746},
  year={2025}
}

@inproceedings{dynamic-3dgaussians,
  title={Dynamic {3D} {Gaussians}: Tracking by persistent dynamic view synthesis},
  author={Luiten, Jonathon and Kopanas, Georgios and Leibe, Bastian and Ramanan, Deva},
  booktitle={3DV},
  pages = {800--809},
  year={2024},
}

@inproceedings{som2024,
  title     = {Shape of Motion: {4D} Reconstruction from a Single Video},
  author    = {Wang, Qianqian and Ye, Vickie and Gao, Hang and Zeng, Weijia and Austin, Jake and Li, Zhengqi and Kanazawa, Angjoo},
  booktitle   = {ICCV},
  pages = {9660--9672},
  year      = {2025}
}

@inproceedings{kplane,
  title={{K-Planes}: Explicit radiance fields in space, time, and appearance},
  author={Fridovich-Keil, Sara and Meanti, Giacomo and Warburg, Frederik Rahb{\ae}k and Recht, Benjamin and Kanazawa, Angjoo},
  booktitle={CVPR},
  pages = {12479--12488},
  year={2023}
}

@inproceedings{zhangmonst3r,
  title={{MonST3R}: A Simple Approach for Estimating Geometry in the Presence of Motion},
  author={Zhang, Junyi and Herrmann, Charles and Hur, Junhwa and Jampani, Varun and Darrell, Trevor and Cole, Forrester and Sun, Deqing and Yang, Ming-Hsuan},
  booktitle={ICLR},
  year={2025}
}

@inproceedings{cut3r,
    Author = {Qianqian Wang and Yifei Zhang and Aleksander Holynski and Alexei A. Efros and Angjoo Kanazawa},
    Title = {Continuous {3D} Perception Model with Persistent State},
    year={2025},
    pages = {10510--10522},
    booktitle={CVPR},
}

@inproceedings{chen2025ttt3r,
    title={{TTT3R}: {3D} Reconstruction as Test-Time Training},
    author={Chen, Xingyu and Chen, Yue and Xiu, Yuliang and Geiger, Andreas and Chen, Anpei},
    booktitle={ICLR},
    year={2026}
    }

@inproceedings{streamVGGT,
    title={Streaming Visual Geometry Transformer}, 
    author={Dong Zhuo and Wenzhao Zheng and Jiahe Guo and Yuqi Wu and Jie Zhou and Jiwen Lu},
    booktitle={ICLR},
    year={2026}
}

@inproceedings{li2025megasam,
  title={{MegaSaM}: Accurate, fast and robust structure and motion from casual dynamic videos},
  author={Li, Zhengqi and Tucker, Richard and Cole, Forrester and Wang, Qianqian and Jin, Linyi and Ye, Vickie and Kanazawa, Angjoo and Holynski, Aleksander and Snavely, Noah},
  booktitle={CVPR},
  pages = {10486--10496},
  year={2025}
}

@inproceedings{reconstructingpeopleplaces,
  title={Reconstructing people, places, and cameras},
  author={M{\"u}ller, Lea and Choi, Hongsuk and Zhang, Anthony and Yi, Brent and Malik, Jitendra and Kanazawa, Angjoo},
  booktitle={CVPR},
  pages = {21948--21958},
  year={2025}
}

@inproceedings{mhmr,
  title={{Multi-HMR}: Multi-person Whole-Body Human Mesh Recovery in a Single Shot},
  author={Baradel, Fabien and Armando, Matthieu and Galaaoui, Salma and Br{\'e}gier, Romain and Weinzaepfel, Philippe and Rogez, Gr{\'e}gory and Lucas, Thomas},
  booktitle={ECCV},
  series = {Lecture Notes in Computer Science},
  volume = {15081},
  pages = {202--218},
  year={2024}
}

@inproceedings{rojas2025hamst3r,
  title={{HAMSt3R}: Human-Aware Multi-view Stereo {3D} Reconstruction},
  author={Rojas, Sara and Armando, Matthieu and Ghanem, Bernard and Weinzaepfel, Philippe and Leroy, Vincent and Rogez, Gregory},
  booktitle={ICCV},
  pages = {5027--5037},
  year={2025}
}

@inproceedings{liu2026josh3r,
    title={Joint Optimization for {4D} Human-Scene Reconstruction in the Wild},
    author={Liu, Zhizheng and Lin, Joe and Wu, Wayne and Zhou, Bolei},
    booktitle={ICLR},
    year={2026}
}

@inproceedings{chen2025human3r,
    title={{Human3R}: Everyone Everywhere All at Once},
    author={Chen, Yue and Chen, Xingyu and Xue, Yuxuan and Chen, Anpei and Xiu, Yuliang and Pons-Moll, Gerard},
    booktitle={ICLR},
    year={2026}
}

@inproceedings{xue2024hsr,
  title={{HSR}: Holistic {3D} Human-Scene Reconstruction from Monocular Videos},
  author={Xue, Lixin and Guo, Chen and Zheng, Chengwei and Wang, Fangjinghua and Jiang, Tianjian and Ho, Hsuan-I and Kaufmann, Manuel and Song, Jie and Hilliges, Otmar},
series = {Lecture Notes in Computer Science},
volume = {15130},
pages = {429--448},
  booktitle={ECCV},
  year={2024},
}

@inproceedings{kocabas2024hugs,
    title={{HUGS}: Human {Gaussian} Splats},
    author={Kocabas, Muhammed and Chang, Jen-Hao Rick and Gabriel, James and Tuzel, Oncel and Ranjan, Anurag},
  booktitle={CVPR},
  pages = {505--515},
  year={2024}
}

@inproceedings{zhang2025odhsr,
  title={{ODHSR}: Online Dense {3D} Reconstruction of Humans and Scenes from Monocular Videos},
  author={Zhang, Zetong and Kaufmann, Manuel and Xue, Lixin and Song, Jie and Oswald, Martin R},
  booktitle={CVPR},
  pages = {21824--21835},
  year={2025}
}

@inproceedings{guo2023vid2avatar,
  title={{Vid2Avatar}: {3D} Avatar Reconstruction from Videos in the Wild via Self-supervised Scene Decomposition},
  author={Guo, Chen and Jiang, Tianjian and Chen, Xu and Song, Jie and Hilliges, Otmar},
  booktitle={CVPR},
  pages = {12858--12868},
  year={2023}
}

@inproceedings{dpt,
  title={Vision transformers for dense prediction},
  author={Ranftl, Ren{\'e} and Bochkovskiy, Alexey and Koltun, Vladlen},
  booktitle={ICCV},
  pages = {12179--12188},
  year={2021}
}

@inproceedings{bedlam,
  title = {{BEDLAM}: A Synthetic Dataset of Bodies Exhibiting Detailed Lifelike Animated Motion},
  author = {Black, Michael J. and Patel, Priyanka and Tesch, Joachim and Yang, Jinlong}, 
  booktitle = {CVPR},
  pages = {8726--8737},
  year = {2023},
}

@inproceedings{lpips,
  title={The unreasonable effectiveness of deep features as a perceptual metric},
  author={Zhang, Richard and Isola, Phillip and Efros, Alexei A and Shechtman, Eli and Wang, Oliver},
  booktitle={CVPR},
  pages = {586--595},
  year={2018}
}

@article{AnySplat,
  title={{AnySplat}: Feed-forward {3D} {Gaussian} Splatting from Unconstrained Views},
  author={Jiang, Lihan and Mao, Yucheng and Xu, Linning and Lu, Tao and Ren, Kerui and Jin, Yichen and Xu, Xudong and Yu, Mulin and Pang, Jiangmiao and Zhao, Feng and Lin, Dahua and Dai, Bo},
  journal={TOG},
  volume={44},
  number={6},
  articleno = {257},
  pages={257:1--257:16},
  year={2025},
}

@inproceedings{depthanything3,
  title={{Depth Anything 3}: Recovering the Visual Space from Any Views},
  author={Haotong Lin and Sili Chen and Jun Hao Liew and Donny Y. Chen and Zhenyu Li and Yang Zhao and Sida Peng and Hengkai Guo and Xiaowei Zhou and Guang Shi and Jiashi Feng and Bingyi Kang},
  booktitle={ICLR},
  year={2026}
}

@inproceedings{yonosplat,
  title={{YoNoSplat}: You Only Need One Model for Feedforward {3D} {Gaussian} Splatting},
  author={Ye, Botao and Chen, Boqi and Xu, Haofei and Barath, Daniel and Pollefeys, Marc},
  booktitle={ICLR},
  year={2026}
}

@inproceedings{co3dv2,
    author={Jeremy Reizenstein and Roman Shapovalov and Philipp Henzler and Luca Sbordone and Patrick Labatut and David Novotny},
    title={Common Objects in {3D}: Large-Scale Learning and Evaluation of Real-life {3D} Category Reconstruction},
    booktitle={ICCV},
    pages = {10901--10911},
    year={2021}
}

@inproceedings{blendedmvs,
    author={Yao Yao and Zixin Luo and Shiwei Li and Jingyang Zhang and Yufan Ren and Lei Zhou and Tian Fang and Long Quan},
    title={{BlendedMVS}: A Large-scale Dataset for Generalized Multi-view Stereo Networks},
    pages = {1790--1799},
    booktitle={CVPR},
    year={2020}
}

@inproceedings{dl3dv,
    author={Lu Ling and Yichen Sheng and Zhi Tu and Wentian Zhao and Cheng Xin and Kun Wan and Lantao Yu and Qianyu Guo and Zixun Yu and Yawen Lu and Xuanmao Li and Xingpeng Sun and Rohan Ashok and Aniruddha Mukherjee and Hao Kang and Xiangrui Kong and Gang Hua and Tianyi Zhang and Bedrich Benes and Aniket Bera},
    title={{DL3DV-10K}: A Large-Scale Scene Dataset for Deep Learning-based {3D} Vision},
    booktitle={CVPR},
    pages = {22160--22169},
    year={2024}
}

@inproceedings{megadepth,
    author={Zhengqi Li and Noah Snavely},
    title={{MegaDepth}: Learning Single-View Depth Prediction from Internet Photos},
    booktitle={CVPR},
    pages = {2041--2050},
    year={2018}
}

@inproceedings{kubric,
    author={Klaus Greff and Francois Belletti and Lucas Beyer and Carl Doersch and Yilun Du and Daniel Duckworth and David J. Fleet and Dan Gnanapragasam and Florian Golemo and Charles Herrmann and Thomas Kipf and Abhijit Kundu and Dmitry Lagun and Issam Laradji and Hsueh-Ti (Derek) Liu and Henning Meyer and Yishu Miao and Derek Nowrouzezahrai and Cengiz Oztireli and Etienne Pot and Noha Radwan and Daniel Rebain and Sara Sabour and Mehdi S. M. Sajjadi and Matan Sela and Vincent Sitzmann and Austin Stone and Deqing Sun and Suhani Vora and Ziyu Wang and Tianhao Wu and Kwang Moo Yi and Fangcheng Zhong and Andrea Tagliasacchi},
    title={Kubric: A scalable dataset generator},
    booktitle={CVPR},
    pages = {3749--3761},
    year={2022}
}

@inproceedings{wildrgb,
  title={{RGBD} Objects in the Wild: Scaling Real-World {3D} Object Learning from {RGB-D} Videos},
  author={Hongchi Xia and Yang Fu and Sifei Liu and Xiaolong Wang},
  booktitle={CVPR},
  pages = {22378--22389},
  year={2024}
}

@inproceedings{scannet,
    author={Angela Dai and Angel X. Chang and Manolis Savva and Maciej Halber and Thomas Funkhouser and Matthias Nießner},
    title={{ScanNet}: Richly-annotated {3D} Reconstructions of Indoor Scenes},
    booktitle={CVPR},
    pages={5828--5839},
    year={2017}
}

@inproceedings{hypersim,
    author={Mike Roberts and Jason Ramapuram and Anurag Ranjan and Atulit Kumar and Miguel Angel Bautista and Nathan Paczan and Russ Webb and Joshua M. Susskind},
    title={Hypersim: A Photorealistic Synthetic Dataset for Holistic Indoor Scene Understanding},
    booktitle={ICCV},
    pages = {10912--10922},
    year={2021}
}

@inproceedings{mapillary,
    author={Manuel Lopez Antequera and Pau Gargallo and Markus Hofinger and Samuel Rota Bulo and Yubin Kuang and Peter Kontschieder},
    title={Mapillary Planet-Scale Depth Dataset},
    booktitle={ECCV},
    series = {Lecture Notes in Computer Science},
    volume = {12347},
    pages = {589--604},
    year={2020}
}

@inproceedings{habitat,
    author={Andrew Szot and Alex Clegg and Eric Undersander and Erik Wijmans and Yili Zhao and John Turner and Noah Maestre and Mustafa Mukadam and Devendra Chaplot and Oleksandr Maksymets and Aaron Gokaslan and Vladimir Vondrus and Sameer Dharur and Franziska Meier and Wojciech Galuba and Angel Chang and Zsolt Kira and Vladlen Koltun and Jitendra Malik and Manolis Savva and Dhruv Batra},
    title={Habitat 2.0: Training Home Assistants to Rearrange their Habitat},
    booktitle={NeurIPS},
volume = {34},
pages = {251--266},
    year={2021}
}

@article{replica,
  title={The {Replica} Dataset: A Digital Replica of Indoor Spaces},
  author={Julian Straub and Thomas Whelan and Lingni Ma and Yufan Chen and Erik Wijmans and Simon Green and Jakob J. Engel and Raul Mur-Artal and Carl Ren and Shobhit Verma and Anton Clarkson and Mingfei Yan and Brian Budge and Yajie Yan and Xiaqing Pan and June Yon and Yuyang Zou and Kimberly Leon and Nigel Carter and Jesus Briales and Tyler Gillingham and Elias Mueggler and Luis Pesqueira and Manolis Savva and Dhruv Batra and Hauke M. Strasdat and Renzo De Nardi and Michael Goesele and Steven Lovegrove and Richard Newcombe},
  journal={arXiv preprint arXiv:1906.05797},
  year={2019}
}

@inproceedings{mvssynth,
    author={Po-Han Huang and Kevin Matzen and Johannes Kopf and Narendra Ahuja and Jia-Bin Huang},
    title={{DeepMVS}: Learning Multi-view Stereopsis},
    booktitle={CVPR},
    pages = {2821--2830},
    year={2018}
}

@inproceedings{pointodyssey,
    author={Yang Zheng and Adam W. Harley and Bokui Shen and Gordon Wetzstein and Leonidas J. Guibas},
    title={{PointOdyssey}: A Large-Scale Synthetic Dataset for Long-Term Point Tracking},
    booktitle={ICCV},
     pages={19855–19865},
    year={2023}
}

@article{virtualkitti,
  title={Virtual {KITTI} 2},
  author={Yohann Cabon and Naila Murray and Martin Humenberger},
  journal={arXiv preprint arXiv:2001.10773},
  year={2020}
}

@inproceedings{aria,
    author={Xiaqing Pan and Nicholas Charron and Yongqian Yang and Scott Peters and Thomas Whelan and Chen Kong and Omkar Parkhi and Richard Newcombe and Carl Yuheng Ren},
    title={Aria Digital Twin: A New Benchmark Dataset for Egocentric {3D} Machine Perception},
    booktitle={ICCV},
    pages = {20133--20143},
    year={2023}
}

@inproceedings{objaverse,
    author={Matt Deitke and Dustin Schwenk and Jordi Salvador and Luca Weihs and Oscar Michel and Eli VanderBilt and Ludwig Schmidt and Kiana Ehsani and Aniruddha Kembhavi and Ali Farhadi},
    title={Objaverse: A Universe of Annotated {3D} Objects},
    booktitle={CVPR},
    pages = {13142--13153},
    year={2023}
}

@inproceedings{NeuMan,
  title={{NeuMan}: Neural Human Radiance Field from a Single Video},
  author={Jiang, Wei and Yi, Kwang Moo and Samei, Golnoosh and Tuzel, Oncel and Ranjan, Anurag},
  booktitle={ECCV},
  series = {Lecture Notes in Computer Science},
  volume = {13692},
  pages = {402--418},
  year={2022},
}

@article{ssim,
  title={Image quality assessment: {From} error visibility to structural similarity},
  author={Wang, Zhou and Bovik, Alan C and Sheikh, Hamid R and Simoncelli, Eero P},
  journal={TIP},
  volume={13},
  number={4},
  pages={600-612},
  year={2004},
}

@inproceedings{emdb,
  author = {Kaufmann, Manuel and Song, Jie and Guo, Chen and Shen, Kaiyue and Jiang, Tianjian and Tang, Chengcheng and Z{\'a}rate, Juan Jos{\'e} and Hilliges, Otmar},
  title = {{EMDB}: The {E}lectromagnetic {D}atabase of {G}lobal 3{D} {H}uman {P}ose and {S}hape in the {W}ild},
  booktitle = {ICCV},
  pages = {14632--14643},
  year = {2023}
}

@inproceedings{qiu2025LHM,
    title={{LHM}: Large Animatable Human Reconstruction Model for Single Image to {3D} in Seconds},
    author={Lingteng Qiu and Xiaodong Gu and Peihao Li  and Qi Zuo
      and Weichao Shen and Junfei Zhang and Kejie Qiu and Weihao Yuan
      and Guanying Chen and Zilong Dong and Liefeng Bo 
      },
    booktitle={ICCV},
    pages = {14184--14194},
  year={2025}
}

@inproceedings{gaussian-reg,
  title={Effective rank analysis and regularization for enhanced {3D} {Gaussian} splatting},
  author={Hyung, Junha and Hong, Susung and Hwang, Sungwon and Lee, Jaeseong and Choo, Jaegul and Kim, Jin-Hwa},
  booktitle={NeurIPS},
  volume = {37},
  pages = {110412--110435},
  year={2024}
}

@article{umeyama2002least,
  title={Least-squares estimation of transformation parameters between two point patterns},
  author={Umeyama, Shinji},
  journal={TPAMI},
  volume={13},
  number={4},
  pages={376-380},
  year={1991},
}

@inproceedings{3dpw,
    title = {Recovering Accurate {3D} Human Pose in The Wild Using {IMUs} and a Moving Camera},
    author = {von Marcard, Timo and Henschel, Roberto and Black, Michael J. and Rosenhahn, Bodo and Pons-Moll, Gerard},
    booktitle = {ECCV},
series = {Lecture Notes in Computer Science},
volume = {11214},
pages = {614--631},
    year = {2018},
}

@inproceedings{adamw,
title={Decoupled Weight Decay Regularization},
author={Ilya Loshchilov and Frank Hutter},
booktitle={ICLR},
year={2019},
}

@inproceedings{chen2024far-nerfmemory,
  title={How far can we compress {Instant-NGP-Based} {NeRF}?},
  author={Chen, Yihang and Wu, Qianyi and Harandi, Mehrtash and Cai, Jianfei},
  booktitle={CVPR},
  pages = {20321--20330},
  year={2024}
}

@inproceedings{li2023steernerf-quicknerf,
  title={{SteerNeRF}: Accelerating {NeRF} Rendering via Smooth Viewpoint Trajectory},
  author={Li, Sicheng and Li, Hao and Wang, Yue and Liao, Yiyi and Yu, Lu},
  booktitle={CVPR},
  pages = {20701--20711},
  year={2023}
}

@inproceedings{movies,
  title={{MoVieS}: Motion-Aware {4D} Dynamic View Synthesis in One Second},
  author={Lin, Chenguo and Lin, Yuchen and Pan, Panwang and Yu, Yifan and Hu, Tao and Yan, Honglei and Fragkiadaki, Katerina and Mu, Yadong},
  booktitle={CVPR},
  pages = {295--306},
  year={2026}
}

@inproceedings{xu20254dgt,
    title     = {{4DGT}: Learning a {4D} {Gaussian} Transformer Using Real-World Monocular Videos},
    author    = {Xu, Zhen and Li, Zhengqin and Dong, Zhao and Zhou, Xiaowei and Newcombe, Richard and Lv, Zhaoyang},
    volume = {38},
    pages = {42606--42632},
    booktitle   = {NeurIPS},
    year      = {2025}
}

@inproceedings{wu2025streamsplat,
  title={{StreamSplat}: Towards Online Dynamic {3D} Reconstruction from Uncalibrated Video Streams},
  author={Wu, Zike and Yan, Qi and Yi, Xuanyu and Wang, Lele and Liao, Renjie},
  booktitle={ICLR},
  year={2026}
}

@inproceedings{multiego,
author = {Li, Bate and Zhong, Houqiang and Cheng, Zhengxue and Hu, Qiang and Wang, Qiang and Song, Li and Zhang, Wenjun},
title = {{MultiEgo}: A Multi-View Egocentric Video Dataset for {4D} Scene Reconstruction},
year = {2025},
booktitle = {ACM MM},
pages = {12882--12889},
doi = {10.1145/3746027.3758232},
}

@inproceedings{d4rt,
  title={Efficiently Reconstructing Dynamic Scenes One {D4RT} at a Time},
  author={Zhang, Chuhan and Le Moing, Guillaume and Koppula, Skanda and Rocco, Ignacio and Momeni, Liliane and Xie, Junyu and Sun, Shuyang and Sukthankar, Rahul and Barral, Jo{\"e}lle K. and Hadsell, Raia and Ghahramani, Zoubin and Zisserman, Andrew and Zhang, Junlin and Sajjadi, Mehdi S. M.},
  booktitle={CVPR},
  pages = {7382--7392},
  year={2026}
}

@InProceedings{warp,
author = {Lai, Wei-Sheng and Huang, Jia-Bin and Wang, Oliver and Shechtman, Eli and Yumer, Ersin and Yang, Ming-Hsuan},
title = {Learning Blind Video Temporal Consistency},
booktitle = {ECCV},
series = {Lecture Notes in Computer Science},
volume = {11219},
pages = {179--195},
year = {2018}
}

@InProceedings{Xiong2024MVHumanNet,
    author    = {Xiong, Zhangyang and Li, Chenghong and Liu, Kenkun and Liao, Hongjie and Hu, Jianqiao and Zhu, Junyi and Ning, Shuliang and Qiu, Lingteng and Wang, Chongjie and Wang, Shijie and Cui, Shuguang and Han, Xiaoguang},
    title     = {{MVHumanNet}: A Large-scale Dataset of Multi-view Daily Dressing Human Captures},
    pages = {19801--19811},
    booktitle = {CVPR},
    year      = {2024},
}

@InProceedings{unish,
    author    = {Li, Mengfei and Li, Peng and Zhang, Zheng and Lu, Jiahao and Zhao, Chengfeng and Xue, Wei and Liu, Qifeng and Peng, Sida and Zhang, Wenxiao and Luo, Wenhan and Liu, Yuan and Guo, Yike},
    title     = {{UniSH}: Unifying Scene and Human Reconstruction in a Feed-Forward Pass},
    booktitle = {CVPR},
    pages = {14038--14049},
    year      = {2026},
}
}

\clearpage
\newpage
\appendix

\section{Implementation Details}
\subsection{Training Details}
\label{subsec:app_training}
We freeze the pretrained Human3R~\cite{chen2025human3r} backbone and train only the Gaussian prediction modules.
For the Scene Gaussian Decoder, we train the DPT-style decoder, the CNN feature decoder and the convolutional head $F_{G}$.
For the Human Gaussian Decoder, we train the Human Gaussian Transformer and the Gaussian MLP $F_{H}$. 
All input images are resized such that the longer side is 512 pixels while preserving the aspect ratio.
We train both decoders using AdamW~\cite{adamw} with a learning rate of $1\times10^{-4}$, weight decay of $1\times10^{-4}$, and a batch size of 2 for the Scene Gaussian Decoder and 1 for the Human Gaussian Decoder.
The Scene Gaussian Decoder is trained for 100k iterations on one NVIDIA A100 80GB GPU, taking approximately 1 day, while the Human Gaussian Decoder is trained for 150k iterations on one NVIDIA A100 40GB GPU, taking approximately two days.
As for the loss function, we set $\lambda_{\mathrm{mse}} = 1.0$, $\lambda_{\mathrm{lpips}} = 0.2$, $\lambda_{\mathrm{reg}} = 0.05$, $\lambda_{\mathrm{dep}} = 0.1$ for $\mathcal{L}_{\mathrm{scene}}$, and $\lambda_{\mathrm{mse}} = 1.0$, $\lambda_{\mathrm{part}} = 0.5$, $\lambda_{\mathrm{reg}} = 100.0$, $\lambda_{\mathrm{sil}} = 1.0$ for $\mathcal{L}_{\mathrm{human}}$.

\subsection{Decomposition-Based Baselines}
\label{subsec:app_baselines}
To evaluate the benefit of jointly reconstructing humans and scenes, we construct decomposition-based baselines by combining existing feed-forward reconstruction methods.
These baselines separately reconstruct the static scenes and each dynamic human, and then compose them in a common coordinate frame using Umeyama's alignment~\cite{umeyama2002least}.

\noindent\textbf{AnySplat.}
As a scene-only baseline, we directly apply AnySplat~\cite{AnySplat} to the input frames.
This baseline reconstructs the scene as static 3D Gaussians and does not explicitly model dynamic humans.

\noindent\textbf{AnySplat+LHM+Human3R.}
We first use Human3R~\cite{chen2025human3r} to estimate human masks, background masks, and SMPL-X~\cite{SMPL-X} parameters.
The background is reconstructed by applying AnySplat only to the masked background regions.
For each detected human, we crop the human region from the input image and reconstruct a canonical human representation using LHM~\cite{qiu2025LHM}.
We then animate and place the reconstructed human using the SMPL-X parameters estimated by Human3R, and compose it with the reconstructed background.
We use the same input and target frames, GT target cameras, and Human3R pose-estimation conditions as our method.
Target frames are excluded from both AnySplat and LHM reconstruction.

\noindent\textbf{AnySplat+LHM+GT Pose.}
We additionally report an oracle variant that replaces the Human3R-estimated masks and SMPL-X parameters with GT annotations.
%

\section{Analysis}
\subsection{Video Depth Estimation}
\label{subsec:exp_depth}
\begin{table}[t]
\centering
\begin{tabular}{lcc}
\toprule
Method & AbsRel $\downarrow$ & $\delta < 1.25 \uparrow$ \\
\midrule
DA3~\cite{depthanything3} & 0.47 & 0.70 \\
Human3R~\cite{chen2025human3r} & 0.54 & 0.71 \\
Ours {\color{red}w/o} depth loss & 0.44 & 0.74 \\
\textbf{Ours} & \textbf{0.43} & \textbf{0.75} \\
\bottomrule
\end{tabular}
\caption{\textbf{Video depth estimation on NeuMan~\cite{NeuMan}.}}
\label{tab:depth}
\end{table}
%
We evaluate geometric accuracy by comparing predicted depth with GT depth using AbsRel and threshold accuracy ($\delta < 1.25$), and compare our performance with prior methods in Table~\ref{tab:depth}.
Following the standard video depth estimation protocol, we estimate a single global scale per sequence and apply it to all frames, instead of performing per-frame scale alignment.
This protocol better reflects dynamic scene reconstruction, as it requires temporally consistent geometry across the video.
Under our sequence-level scale alignment protocol, our method achieves lower AbsRel and higher threshold accuracy than DA3~\cite{depthanything3} and Human3R~\cite{chen2025human3r}.

\subsection{Static Scene Reconstruction}
\label{subsec:exp_static}
\begin{table}[t]
\centering
\renewcommand{\arraystretch}{1.12}
\setlength{\tabcolsep}{3pt}
\resizebox{\linewidth}{!}{%
\begin{tabular}{l cc ccc ccc ccc ccc}
\toprule
\textbf{Method}
& \textbf{FF}
& \textbf{Str.}
& \multicolumn{3}{c}{\textbf{NeuMan~\cite{NeuMan} (4-view)}}
& \multicolumn{3}{c}{\textbf{NeuMan~\cite{NeuMan} (16-view)}}
& \multicolumn{3}{c}{\textbf{EMDB~\cite{emdb} (4-view)}}
& \multicolumn{3}{c}{\textbf{EMDB~\cite{emdb} (16-view)}} \\
\cmidrule(lr){4-6}
\cmidrule(lr){7-9}
\cmidrule(lr){10-12}
\cmidrule(lr){13-15}
& & 
& PSNR$\uparrow$ & SSIM$\uparrow$ & LPIPS$\downarrow$
& PSNR$\uparrow$ & SSIM$\uparrow$ & LPIPS$\downarrow$
& PSNR$\uparrow$ & SSIM$\uparrow$ & LPIPS$\downarrow$
& PSNR$\uparrow$ & SSIM$\uparrow$ & LPIPS$\downarrow$ \\
\midrule

HSR~\cite{xue2024hsr}
& {\color{red}\ding{55}} & {\color{red}\ding{55}}
& \underline{22.1} & \underline{0.62} & 0.54
& \underline{19.7} & \underline{0.62} & 0.55
& \textbf{21.5} & \textbf{0.71} & 0.45
& 17.3 & \textbf{0.73} & 0.45 \\

AnySplat~\cite{AnySplat}
& {\color{Green}\ding{51}} & {\color{red}\ding{55}}
& \textbf{24.0} & \textbf{0.82} & \textbf{0.12}
& \textbf{20.6} & \textbf{0.65} & \textbf{0.24}
& \underline{19.8} & 0.67 & \textbf{0.24}
& 17.8 & 0.62 & \underline{0.34} \\

YoNoSplat~\cite{yonosplat}
& {\color{Green}\ding{51}} & {\color{red}\ding{55}}
& 14.7 & 0.44 & 0.49
& 16.5 & 0.47 & 0.50
& 14.4 & 0.48 & 0.52
& 16.2 & 0.58 & 0.47  \\

DA3~\cite{depthanything3}
& {\color{Green}\ding{51}} & {\color{red}\ding{55}}
& 20.7 & 0.57 & \underline{0.27}
& 18.5 & 0.53 & \underline{0.38}
& 19.4 & 0.61 & \underline{0.28}
& \textbf{19.1} & 0.65 & \textbf{0.32} \\

\textbf{Ours}
& {\color{Green}\ding{51}} & {\color{Green}\ding{51}}
& 19.7 & 0.60 & 0.26
& 17.4 & 0.47 & 0.39
& \underline{19.8} & \underline{0.68} & \underline{0.28}
& \underline{18.8} & \underline{0.66} & 0.35 \\

\bottomrule
\end{tabular}%
}
\caption{
\textbf{Single-human scene novel view synthesis on NeuMan~\cite{NeuMan} and EMDB~\cite{emdb}.}
We report results on \textit{background regions only}.
\textbf{FF} and \textbf{Str.} denote feed-forward and streaming inference, respectively.
}
\label{tab:background}
\end{table}
We additionally evaluate novel view synthesis on static background regions to isolate the scene reconstruction capability of each method.
As shown in Table~\ref{tab:background}, HSR~\cite{xue2024hsr} achieves higher PSNR and SSIM in several settings, benefiting from per-scene optimization over the target sequence.
However, our method obtains better LPIPS, suggesting that it preserves perceptual scene quality despite operating in a feed-forward manner.
On background-only metrics, our method performs worse than AnySplat~\cite{AnySplat} on NeuMan, while its performance relative to AnySplat on EMDB and to DA3~\cite{depthanything3} on both datasets varies across metrics and input-view settings.
%

%
AnySplat and DA3 process input frames jointly, whereas our method reconstructs the scene sequentially while modeling dynamic humans. This difference in input processing may contribute to the observed performance differences.

\subsection{Long-Term Streaming Scalability}
\label{subsec: streaming_scalability}
We evaluate the long-term scalability of our representation on a 500-frame sequence.
To isolate the effect of background aggregation, we process the frames with voxel edge lengths of $0.01$, $0.03$, and $0.05$\,m.

\begin{figure}[t]
    \centering
    \begin{subfigure}[t]{0.49\textwidth}
        \centering
        \includegraphics[width=\linewidth]{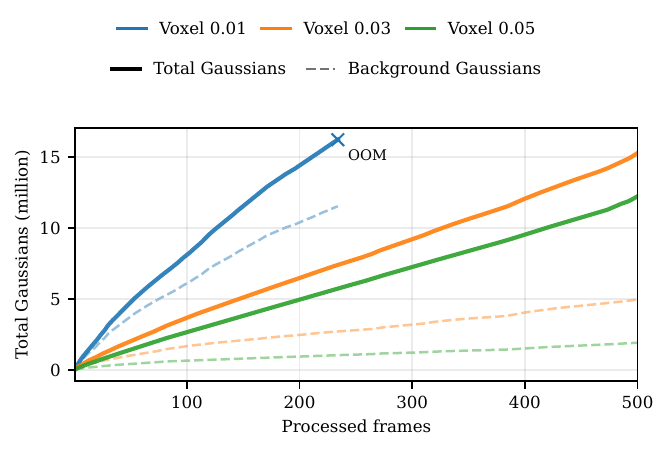}
        \caption{Accumulated Gaussian count. Solid curves show the total Gaussian count, while the lighter dashed curves show the background-only count. The cross indicates where the voxel size of $0.01$ runs out of GPU memory.
        }
        \label{fig:streaming_gaussians}
    \end{subfigure}
    \hfill
    \begin{subfigure}[t]{0.49\textwidth}
        \centering
        \includegraphics[width=\linewidth]
        {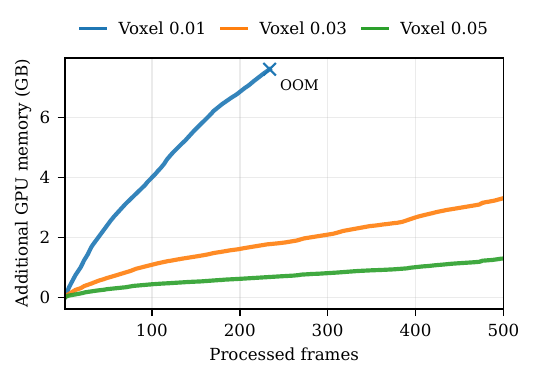}
        \caption{Additional allocated GPU memory relative to the first frame. 
        }
        \label{fig:streaming_memory}
    \end{subfigure}
    \caption{
    Long-term streaming scalability over 500 frames with different background
    voxel sizes. All configurations process the same input frames with
    identical model and inference settings.
    }
    \label{fig:streaming_scalability}
\end{figure}

Figure~\ref{fig:streaming_gaussians} analyzes the growth of the Gaussian representation.
The total count includes the retained background Gaussians and all human Gaussians generated up to each frame.
Figure~\ref{fig:streaming_memory} shows the corresponding GPU memory consumption.
Background Gaussians occupying the same spatial voxel are merged, whereas human Gaussians are accumulated across frames to represent the dynamic human sequence.

With a voxel size of $0.01$\,m, the representation reaches 16.20 million total Gaussians and runs out of GPU memory after 234 frames.
In contrast, voxel sizes of $0.03$\,m and $0.05$\,m successfully process all 500 frames, resulting in 15.26
million and 12.23 million total Gaussians, respectively.
The corresponding additional GPU memory consumption is 3.31 GB and 1.30 GB.
In particular, the background-only curves show that larger voxels substantially suppress the
growth of the persistent background representation.
These results demonstrate the trade-off between spatial granularity and long-term streaming scalability.

\subsection{Runtime Breakdown}
\label{subsec: runtime_breakdown}
We decompose the inference time into Human3R inference, background Gaussian decoding, human Gaussian decoding, and update and I/O operations.
Table~\ref{tab:runtime_breakdown} reports the average runtime over the NeuMan~\cite{NeuMan} evaluation sequences with eight input frames.
\begin{table}[t]
\centering
\caption{Runtime breakdown of our method on NeuMan.}
\label{tab:runtime_breakdown}
\resizebox{\linewidth}{!}{%
\begin{tabular}{lcc}
\toprule
Component & Time (ms/frame) & Ratio (\%) \\
\midrule
Human3R inference           & 264.82 & 49.0 \\
Scene Gaussian Decoder      &  73.05 & 13.5 \\
Human Gaussian Decoder      &  39.47 &  7.3 \\
Update and I/O              & 162.93 & 30.2 \\
\midrule
Total                       & 540.27 & 100.0 \\
\bottomrule
\end{tabular}
}
\end{table}
Human3R inference is the primary computational bottleneck, accounting for 49.0\% of the total runtime.
In contrast, the background and human Gaussian decoders together account for only 20.8\%.
This indicates that the main runtime cost originates from the underlying recurrent reconstruction
backbone rather than the proposed Gaussian prediction modules.
Therefore, future acceleration of the Human3R backbone, especially its encoding procedure, would directly improve the overall processing speed.

\section{Limitations}
\label{sec:limitation}
Although our method enables feed-forward photorealistic reconstruction of dynamic human-scene representations, several limitations remain.
First, it relies on geometric and human priors from the underlying foundation model~\cite{chen2025human3r}; therefore, the errors in camera estimation, scene point maps, human detection, or SMPL-X fitting can propagate to the final reconstruction.
Second, severe human-human occlusions and complex interactions remain challenging, as monocular videos provide limited cues for reliable identity association and complete human geometry.
Finally, visible rendering artifacts remain, particularly in faces, hands, and clothing under motion blur, large pose changes, or limited observations.
Assigning one Gaussian per SMPL-X vertex may further limit the representation of fine-scale geometry, such as hair and loose clothing.

\section{Supplementary Video}
The supplementary video provides additional qualitative results complementing the figures and quantitative evaluations in the paper.
It first presents the input monocular videos alongside our reconstructed dynamic human-scene representations rendered from multiple viewpoints, illustrating the reconstructed 3D structure and temporal consistency.
It then provides side-by-side novel-view video and overview comparisons among the ground truth, our method, StreamSplat~\cite{wu2025streamsplat}, and MoVieS~\cite{movies}, highlighting differences in human appearance, scene geometry, and temporal stability.

\end{document}